\documentclass[a4paper,fleqn]{cas-sc}
\usepackage[authoryear]{natbib}

\usepackage{graphicx}
\usepackage{amsmath,amssymb}
\usepackage[geometry]{ifsym}
\usepackage{mathtools}
\usepackage{autobreak}
\usepackage{amsfonts}
\usepackage{bbm}
\usepackage{colortbl}
\usepackage{url}
\usepackage[subrefformat=parens]{subcaption}
\usepackage{comment}
\usepackage{CJKutf8}
\usepackage{hyperref}
\usepackage{multirow}
\usepackage{booktabs}
\usepackage{array, makecell}
\usepackage{boldline}
\usepackage{tabularx,lipsum}
\usepackage{color}
\usepackage{xcolor}
\usepackage{lastpage}

\usepackage{fancyhdr}

\newcommand{\modified}[1]{\textcolor{black}{#1}}
\newcommand{\reviewera}[1]{\textcolor{black}{#1}}
\newcommand{\reviewerb}[1]{\textcolor{black}{#1}}
\newcommand{\figref}[1]{Figure \ref{#1}}
\renewcommand{\tabref}[1]{Table \ref{#1}}
\newcommand{\secref}[1]{Section \ref{#1}}
\newcommand{\blueblack}{011959}
\newcommand{\thinpink}{FACCFA}
\newcommand{\dirtyyellow}{818232}
\newcommand{\colorboxself}[1]{\fcolorbox[HTML]{#1}{#1}{\rule{0pt}{3pt}\rule{3pt}{0pt}}}
\newcommand{\urlref}[1]{\href{#1}{#1}}

\begin{document}
\let\WriteBookmarks\relax
\def\floatpagepagefraction{1}
\def\textpagefraction{.001}

\shorttitle{Speciesist Language and Nonhuman Animal Bias in English Masked Language Models}

\shortauthors{Takeshita et~al.}

\title [mode = title]{Speciesist Language and Nonhuman Animal Bias \\ in English Masked Language Models}                      



%
\author[1]{Masashi Takeshita}[type=editor,
                        auid=000,bioid=1,
                        orcid=0000-0001-5853-3262]

\cormark[1]


\ead{takeshita.masashi@ist.hokudai.ac.jp}


\credit{Conceptualization of this study, Methodology, Software, Investigation, Data curation, Writing - Original Draft, Visualization}

\address[1]{Graduate School of Information Science and Technology, Hokkaido University}

\author[2]{Rafal Rzepka}[
orcid=0000-0002-8274-0875]
\ead{rzepka@ist.hokudai.ac.jp}

\credit{Writing - Review \& Editing, Supervision}

\author[2]{Kenji Araki}[%
   orcid=0000-0001-9668-1610,
   ]
\ead{araki@ist.hokudai.ac.jp}

\credit{Supervision}

\address[2]{Faculty of Information Science and Technology, Hokkaido University}

\cortext[cor1]{Corresponding author}



\begin{abstract}
Warning: This paper contains examples of offensive language, including insulting or objectifying expressions.

Various existing studies have analyzed what social biases are inherited by NLP models.
These biases may directly or indirectly harm people, therefore previous studies have focused only on human attributes.
However, until recently no research on social biases in NLP regarding nonhumans existed.
In this paper\footnote{Anonymous previous version of this paper is accessible at \urlref{https://openreview.net/forum?id=dfqMpjZOgv4}, we have also published a paper on this topic in Japanese \citep{takeshita-2021-english-speciesist-bias}}, we analyze biases to nonhuman animals, i.e. speciesist bias, inherent in English Masked Language Models \modified{such as BERT}.
\modified{We analyzed speciesist bias against 46 animal names using template-based and corpus-extracted sentences containing speciesist (or non-speciesist) language.
We found that pre-trained masked language models tend to associate harmful words with nonhuman animals and have a bias toward using speciesist language for some nonhuman animal names.}
Our code for reproducing the experiments will be made available on GitHub\footnote{\urlref{https://github.com/Language-Media-Lab/speciesist-language}}.

\end{abstract}




\maketitle
\renewcommand{\headrulewidth}{0.0pt}
\renewcommand{\footrulewidth}{0.5pt}
\thispagestyle{fancy}
\pagestyle{fancy}
    \lhead{} 
    \chead{Speciesist Language and Nonhuman Animal Bias in English Masked Language Models} 
    \rhead{} 
    \lfoot{Takeshita et al.: \textit{Accepted Manuscript}} 
    \cfoot{} 
    \rfoot{\thepage} 

\section{Introduction}

Recently, in the field of Natural Language Processing (NLP), Masked Language Models (MLMs) using Transformers \citep{vaswani2017attention}, such as BERT \citep{devlin-etal-2019-bert} and RoBERTa \citep{liu2019roberta}, widely contributed to the state-of-the-art methods in downstream tasks.
However, existing studies suggest that these models inherit social biases~\citep{Sun2019,Blodgett2020}.
\reviewerb{Here, social bias is defined as the systematic, unequal treatment based on one’s social attribute~\citep[cf.][]{stanczak2021survey-gender-bias-nlp,Sun2019}.}
Such biases cause differences in accuracy between majority and minority attributes  \citep[e.g.][]{romanov-etal-2019-whats} and negative generalizations, e.g. in text generation \citep{liu-etal-2020-does-gender,sheng-etal-2019-woman-worked-as-language-generation,sheng-etal-2021-societal-generation-survey,garimella-etal-2021-he-is-very}.

The studies of social bias in NLP target gender \citep[e.g.][]{Bolukbasi2016,Caliskan2017}, race \citep[e.g.][]{manzini-etal-2019-black}, religion and ethnicity \citep[e.g.][]{li-etal-2020-unqovering} and so on, all of which assume human attributes. 
However, to the best of the authors' knowledge, there are no similar bias studies on \textbf{nonhuman animals}.

In this paper, we use templates, corpus-extracted sentences and pre-trained MLMs to investigate if the bias regarding nonhuman, i.e. \textbf{speciesist bias}, is inherent in MLMs trained on English corpora.

The bias we investigate in this paper is the \textbf{representational bias}, following the classification of \citet{Sun2019} and \citet{Blodgett2020}.
Currently, nonhuman animals do not use the NLP system directly, so we do not need to consider the idea of, e.g. ``performance against the social group of nonhuman animals''. 
On the other hand, we think that we should respect nonhuman animals \textbf{for their own sake, not for the sake of humans} \citep[cf.][]{owe2021moral-nonhumans-ethics-inai}, for the reasons described below, and therefore we should study, for example, insulting associations with nonhuman animals and negative stereotyping against them.

The structure of this paper is as follows. 
This section will be followed by an explanation of the reasons for conducting bias studies on nonhuman animals, in response to possible criticisms.
In \secref{sec:related-work}, we explain related works -- we begin with an overview of social bias research in \modified{pre-trained} language models, then move on to speciesism and speciesist language.
We will describe the experimental setting in \secref{sec:experimental-setup} and the experimental methods in \secref{sec:bias-evaluation-method}. 
In \secref{sec:results}, we report experimental results. 
In \secref{sec:discussion} we discuss these results and the limitations of the experiments presented in this paper. 
\modified{In \secref{sec:mitigate-bias}, given experimental results, we discuss how to mitigate speciesist bias.}
Finally, we conclude our paper in \secref{sec:conclusion}.

\subsection{\reviewera{Ethical Discussion: Nonhumans and NLP}}
\label{subsec:ethical-discussion}
There may be two possible criticisms of the research objectives of this paper. 
The first criticism is that there is no ethical problem with the existence of harmful bias to nonhuman animals. 
However, we should give equal consideration to interests and should not discriminate based on \textbf{who} has the interests \citep{Singer2009-animal-liberation}.  
Even if one does not accept this idea, most people would agree that nonhuman animals deserve some moral consideration \citep{owe2021moral-nonhumans-ethics-inai}\footnote{You can see the more argument about moral status of about nonhuman animals in \citet{regan2004case-for-animal-rights}, \citet{hagendorff2022speciesist-bias} or textbooks of animal ethics  \citep[e.g.][ch.2]{Fischer2021-animal-ethics-introduction}.}. 
If this is true, then it is important to study the biases that are harmful to nonhuman animals.

The second potential criticism is that even if nonhuman animals deserve some moral consideration, NLP models with speciesist bias do not harm them because they do not use it directly.
However, we think it is important to study the speciesist bias of NLP models for three following reasons.

First, if NLP systems with a speciesist bias are popularized in our society, the bias of the NLP system may affect us and thereby indirectly harm animals \citep[in human animal cases, see][]{emily-etal-2021-danger-stochastic-parrots}\footnote{``Stochastic parrots'' in the title of \citet{emily-etal-2021-danger-stochastic-parrots} is an example of specisist language use.}.
For example, if an NLP system generates speciesist sentences, the speciesist bias may propagate to readers who read the sentences, and they may acquire an implicit discriminatory bias against nonhuman animals\footnote{\citet{menegatti2017gender-bias-in-language} argue that sexist language affects our perception, e.g., adversely has impact on women. We think that speciesist language can have similar effect.}.
As we discuss in \secref{subsec:speciesism-and-language}, most people are already discriminatory against nonhuman animals, but 
this phenomenon should not be reinforced.
\modified{In addition, it may be harmful not only to nonhuman animals, but also to those who oppose speciesism and use non-speciesist language. For example, if the text generated is such that nonhuman animals are represented as foods, vegetarians (vegans) who oppose speciesism will be offended and will not use the technology. This will lead to the exclusion of minorities from the technology~\citep[cf.][]{Blodgett2020}. If one considers the exclusion from technology of vegetarians or vegans as unimportant, it is discrimination against them~\citep{horta2018discrimination-vegans}.}

Second, the representational speciesist bias should be considered unwarranted in itself, even if it does no direct harm \citep{Blodgett2020}.
The use of language that is insulting to or demeaning nonhuman animals, as described in \secref{subsec:speciesism-and-language}, is wrong in itself \citep[cf.][]{hellman2008discrimination}, even if nonhuman animals never recognize the expression.

Third, the biases inherent in word embeddings reflect social biases which exist in our cognition, beliefs and social structures \citep{Caliskan2017,garg2018word-embedding-100,joseph-morgan-2020-word-reflect-belief}. 
Therefore, analyzing the speciesist bias in word embeddings and corpora can contribute to research about the influence of this bias on our cognition and society.

For these reasons, we think that it is important to study the speciesist bias in NLP.
\section{Related Work}
\label{sec:related-work}



\subsection{Social Bias in \modified{Pre-trained} Language Models}

Since the report by \citet{Bolukbasi2016} of an inherent gender bias in word embedding, various social biases have been reported to be inherent in static and \reviewera{contextualized} word embedding\footnote{For a survey of bias assessment methods, see \citet{dev2021-what-bias-measure}.}.
Previous studies have typically focused on binary gender and race in bias research \citep[especially,][]{Bolukbasi2016,Caliskan2017,manzini-etal-2019-black}, \citet{nangia-etal-2020-crows-pairs}, for example, they have created bias assessment datasets for as many as nine different attributes.
However, these are all bias studies \textbf{only on the human animal}, and to the best of the authors' knowledge, until recently there \reviewera{were} \textbf{no bias studies on nonhuman animals}.

A notable exception is the very recent study by \citet{hagendorff2022speciesist-bias}. They have \reviewera{conducted} research on speciesist biases in image recognition,
recommendation systems, and analyzed static word embedding (word2vec \citep{Mikolov2013}, GloVe \citep{pennington2014glove}), GPT-3 \citep{brown2020language-few-shot-leaner-gpt-3} and Delphi \citep{jiang2021delphi} by using specific word lists and template sentences.
There are two differences between their study and ours.
First, our experiments are not limited to a specific word list, and we also conduct analysis without template sentences (\secref{subsec:corpus-experiment}).
Second, we are analyzing speciesist bias in relation to speciesist language.
We hope that this new topic is the beginning of a series of research endeavours for analyzing and mitigating speciesist bias.

Bias assessment methods include intrinsic and extrinsic assessment \citep{goldfarb-tarrant-etal-2021-intrinsic-bias}.
Intrinsic evaluation evaluates the bias in the word embedding space itself, while extrinsic one evaluates the bias in the downstream \modified{tasks}.
Since there are currently no downstream tasks in which nonhuman animals are direct users, we use intrinsic evaluation in this paper.
However, we think that extrinsic evaluation is necessary for tasks such as text generation, which is a downstream task but whose effects on unexpected users must be considered.

There are three methods of intrinsic evaluation of bias for \reviewera{contextualised} word embeddings, especially MLMs: using \textbf{template sentences} \citep{bartl-etal-2020-unmasking,hutchinson-etal-2020-social-barriers,kurita-etal-2019-measuring,may-etal-2019-measuring-sentence-SEAT,TanandCelis-2019-assessing-social-intersectional,webster2020measuring-pretrained-models,silva-etal-2021-towards-comprehensive-socialbias-PLMs,Lauscher_etal_2020-general-framework-implicit-explicit-debias, Dev_Li_Phillips_Srikumar_2020-on-measuring-mitigating-bias-NLI,dev-etal-2021-oscar}, \textbf{corpus sentences} \citep{zhao-etal-2019-gender-contextualizedWE,bordia-bowman-2019-identifying-reducing-gender-bias,basta-etal-2019-evaluating,guo2020detecting-CEAT}, and \textbf{non-template sentences created manually} \citep{nadeem-etal-2021-stereoset,nangia-etal-2020-crows-pairs}.
The method using template sentences allows the analysis to focus on specific aspects of bias and is easy to evaluate once the template sentences are created.
In the case of using sentences extracted from a corpus, it is possible to evaluate the bias using actual sentences, which allows for a more realistic experimental setting.
The method of using manually created non-template sentences is more costly to create, but allows for evaluation in a wider range of realistic settings than using template sentences, and also helps to evaluate the bias in more appropriately limited settings than using sentences extracted from a corpus.

As we will discuss in detail in \secref{sec:bias-evaluation-method}, due to the cost of manually creating sentences and other obstacles inherent in this research, to evaluate speciesist bias we use template sentences and sentences extracted from corpora.

\subsection{Speciesism and Language}
\label{subsec:speciesism-and-language}

Speciesism is ``the unjustified comparatively worse consideration or treatment of those who do not belong to a certain species.'' \citep[p.3]{horta-and-akbersmeier-2020-Defining-speciesism}. 
Nonhuman animals, as sentient beings, deserve equal consideration with human animals \citep[p.40]{Singer2009-animal-liberation}, and we should not discriminate against nonhuman animals.
However, we do so, for example by eating their flesh or conducting experiments on them \citep[ch.2, 3]{Singer2009-animal-liberation}.

We also treat nonhuman animals as inferior beings or objects in our language use. 
For instance, ``terming a woman a `dog''' 
 insults all women indirectly and also insults all dogs directly \citep[p.12]{dunayer1995sexist}. 
Usual referring to nonhuman animals as “it” or “something,” or using “that” or “which” as relative pronouns to indicate nonhuman animals are examples of treating nonhuman animals as objects \citep{dunayer2001animal-equality,dunayer_2003_english_and_spe}.
\citet[ch.9]{dunayer2001animal-equality} also states that, in the process of slaughtering, people use  words such as ``harvest'', ``package'' and ``process'' to hide cruelty.
The use of such language and words is referred to as \textbf{speciesist language}, in analogy with sexist language and racist language.
\reviewerb{Non-speciesist language is the use of language that does not treat nonhuman animals as objects or as inferior to human animals, such as using ``who'' to refer nonhuman animals, instead of using ``which''.}

In addition to research conducted in Animal Ethics field, there are also studies in Corpus Linguistics that analyzed language use regarding nonhuman animals.
\citet{jepson2008linguistic-analysis-killing} performed discourse analysis on various texts and spoken conversations showing that the word ``slaughter'' in human context collocates strongly with negative emotions, but lacks such sentiment when used in the context of nonhuman animals.
\citet{franklin2020acts-phd-corpus-linguistics} also analyzed the use of ``killing'' terms, such as ``kill'' and ``slaughter'', in ``People, Products, Pests and Pets'' (PPPP)\footnote{\urlref{https://animaldiscourse.wordpress.com/}} which is an English corpus that contains texts referring to nonhuman animals extracted from various domains such as food-related websites and news articles \citep{Sealey-2018-first-catch-PPPP-corpus}.

Existing studies have reported that stylistic biases are reflected in NLP models \citep{tan-etal-2020-morphin-time,hovy-etal-2020-youshouldsound-likeyourfather}. 
Therefore, since the above-mentioned speciesist language and biases in English may be reflected in MLMs, we investigate a possibility of speciesist bias in English MLMs.


\begin{table}[tb]
    \centering
    \caption{Animal names used in this research and their frequencies in English Wikipedia. The coloring of animal names was done by the authors: \colorboxself{\thinpink} refers to ``farm'' animals, \colorboxself{\dirtyyellow} represents popular nonhuman companions and \colorboxself{\blueblack} addresses all remaining species.}
    \small
    \scalebox{1}{
    \begin{tabular}{|l|r||l|r|}
    \hline
        Animal name & Frequency & Animal name & Frequency \\ \hline\hline
        \colorboxself{\blueblack} horse  & 194,363 & \colorboxself{\blueblack} deer  & 43,130 \\ \hline
      \colorboxself{\thinpink} turkey & 187,079 &\colorboxself{\blueblack} seal & 42,533 \\ \hline
         \colorboxself{\blueblack} fox & 176,569 &  \colorboxself{\blueblack} snake& 42,323 \\ \hline
         \colorboxself{\blueblack} human& 173,145 &  \colorboxself{\dirtyyellow} persian& 39,764 \\ \hline
         \colorboxself{\blueblack} fish& 142,508 &  \colorboxself{\thinpink} duck& 36,828 \\ \hline
         \colorboxself{\dirtyyellow} dog & 127,775 &  \colorboxself{\blueblack} swan& 36,556 \\ \hline
         \colorboxself{\blueblack} bird& 124,463 &  \colorboxself{\thinpink} sheep & 34,433 \\ \hline
         \colorboxself{\blueblack} moth& 93,670 &  \colorboxself{\thinpink} chicken & 34,231 \\ \hline
        \colorboxself{\blueblack} buffalo & 91,392 &  \colorboxself{\blueblack} snail & 33,725 \\ \hline
         \colorboxself{\blueblack} robin & 89,168 &  \colorboxself{\dirtyyellow} bombay & 32,819 \\ \hline
         \colorboxself{\dirtyyellow} cat & 83,038 &  \colorboxself{\blueblack} frog& 31,922 \\ \hline
         \colorboxself{\blueblack} wolf& 78,795 &  \colorboxself{\blueblack} crane& 31,328 \\ \hline
         \colorboxself{\blueblack} eagle & 78,126 &  \colorboxself{\blueblack} penguin& 30,769 \\ \hline
         \colorboxself{\blueblack} bear& 69,029 &  \colorboxself{\blueblack} rat& 28,851 \\ \hline
         \colorboxself{\blueblack} lion& 67,774 &  \colorboxself{\blueblack} monkey& 28,144 \\ \hline
         \colorboxself{\blueblack} tiger& 60,709 &  \colorboxself{\blueblack} falcon& 27,843 \\ \hline
         \colorboxself{\blueblack} beetle& 54,887 &  \colorboxself{\blueblack} rabbit& 27,039 \\ \hline
         \colorboxself{\blueblack} bat& 49,445 &  \colorboxself{\blueblack} beaver& 26,421 \\ \hline
         \colorboxself{\blueblack} mouse& 48,866 &  \colorboxself{\blueblack} pike& 25,392 \\ \hline
         \colorboxself{\blueblack} fly& 45,411 &  \colorboxself{\thinpink} pig& 25,273 \\ \hline
         \colorboxself{\dirtyyellow} \footnotesize{newfoundland}& 44,353 &  \colorboxself{\blueblack} elephant& 24,817 \\ \hline
         \colorboxself{\blueblack} tang& 44,245 &  \colorboxself{\thinpink} cow& 22,563 \\ \hline
         \colorboxself{\blueblack} butterfly& 44,096 &  \colorboxself{\blueblack} molly& 21,353 \\ \hline
    \end{tabular}
    }
    \label{tab:animal_name_freq_list}
\end{table}

\section{Experimental setup}
\label{sec:experimental-setup}

The MLMs used in this paper are BERT$_{\rm LARGE\text{-}cased}$\footnote{ \urlref{https://huggingface.co/bert-large-cased}}, RoBERTa$_{\rm LARGE}$\footnote{\urlref{https://huggingface.co/roberta-large}}, DistilBERT$_{\rm base\text{-}cased}$\footnote{\urlref{https://huggingface.co/distilbert-base-cased}} \citep{sanh2019distilbert} and ALBERT$_{\rm large\text{-}v2}$\footnote{\urlref{https://huggingface.co/albert-large-v2}} \citep{Lan2020ALBERT}, which are widely used in current NLP.
\reviewerb{BERT, DistilBERT and ALBERT were pre-trained on Wikipedia and BookCorpus~\citep{Zhu-2015-bookscorpus}. RoBERTa was pre-trained on Wikipedia, BookCorpus, CC-NEWS, OpenWebText~\citep{Gokaslan2019OpenWeb} and STORIES~\citep{trinh2018simple-commonsense-reasoning-stories}. DistilBERT and ALBERT are smaller models than BERT and RoBERTa. Model details are shown on \tabref{tab:model-details} in Appendix \ref{sec:appendix}.}

We determine animals we focus on in this paper as follows:
\begin{enumerate}
  \setlength{\parskip}{.05cm} 
  \setlength{\itemsep}{0cm} 
    \item We collect animal names from ``All Animals A-Z List.''\footnote{\urlref{https://a-z-animals.com/animals/}} We focus on only one-term names.
    \item We limited the number of animals for this research by choosing only these which names appear on English Wikipedia\footnote{We use the Wikipedia dataset downloaded on 01/05/2020 from \urlref{https://huggingface.co/datasets/wikipedia}.} more than 20,000 times, resulting in 46 animal names in total.
\end{enumerate}
Our hypothesis is that if MLMs recognize different animals by categorizing them, then similar bias will be found for animals in similar contexts.
In this paper, we categorize animals who live in farms to be utilized as flesh marking them in \colorboxself{\thinpink}, nonhuman companions in \colorboxself{\dirtyyellow}, and other animals in \colorboxself{\blueblack} colors, respectively\footnote{Following \citet{crameri2020misuse-colour}, in this paper we use scientific color map \citep{crameri_fabio_scientific_colour_map} to include people with diverse color vision.}. 
In \tabref{tab:animal_name_freq_list}, we show all animal names under investigation, their corresponding colors, and their frequencies in Wikipedia.

As shown in \tabref{tab:animal_name_freq_list}, the taxonomic classes of the animal names used in this experiment are not the same. For example, ``dog'' is a taxonomic subspecies, while ``bear'' is a family.
Initially, we attempted to standardize the taxonomic classes, but in doing so, we had to exclude some animal names that are used in everyday life. 
In the first place, we see no special reason to keep the same taxonomic hierarchy, since the extension of animal names as indicated in everyday terminology may not coincide with the extension of animal names in scientific taxonomy \citep[ch.5, 10]{yoon2009naming-nature}.
In this experiment, we \modified{therefore} decided not to standardize the taxonomic classes because this would be undesirable from the viewpoint of bias evaluation.

In addition, some words such as ``turkey'' and ``crane'' have senses other than animal names, and it is possible that MLMs express these meanings in their internal parameters. However, we do not exclude such animal names from our experiments because we do not know a priori what meanings and biases are reflected in them.


\section{Bias Analysis by Speciesist and Non-Speciesist Language}
\label{sec:bias-evaluation-method}

In this section, we explain the method for evaluating bias in MLMs.
In this experiment, we want to evaluate the relationship between speciesist language and speciesist bias, thus we assess the speciesist bias of MLMs by changing between speciesist and non-speciesist languages.
There are \modified{two more} reasons for using this method. 
First, there is no list of biased words for evaluating bias against nonhuman animals, so we need to evaluate the bias in a way that does not rely on a pre-defined word list.
Second, while most people are aware of gender or racial bias, we think that only few people  consider the speciesist bias to be problematic. Therefore, it would be difficult to hire annotators to create a dataset for speciesist bias evaluation.
For example, the word ``slaughter'' is considered to be a biased word for nonhuman animals. However, since most people would not consider this word to be problematic in the context of meat processing \citep{jepson2008linguistic-analysis-killing}, there is a high probability that annotators would not include this word in the evaluation word list.

In this experiment, we evaluate the bias using two approaches: the template-based and the corpus-based.
The template-based approach is commonly used in bias analysis studies of NLP models, and we believe it is useful for discovering specific biases \citep{kurita-etal-2019-measuring,bartl-etal-2020-unmasking,webster2020measuring-pretrained-models}.
However, this approach may limit the bias that can be evaluated depending on the template prepared \citep{guo2020detecting-CEAT}.
For this reason, we use raw sentences extracted from the corpus.

In \secref{subsec:template-based}, we describe the template-based experiment.
The template sentences used in this experiment are introduced first, and then two experiments using these sentences are described (\secref{subsec:bias-evaluation-probdiff} and \secref{subsec:bias-eval-sent-analysis}, respectively).
Finally, in \secref{subsec:corpus-experiment}, we explain the corpus-based experiment.

\subsection{Template-based Experiment}
\label{subsec:template-based}

The basic template sentence we utilize is \textbf{``[PRONOUN] is a [ANIMAL] [REL-PRONOUN] is [MASK].''}, where [PRONOUN] slot indicates a pronoun, [ANIMAL] is an animal name, and [REL-PRONOUN] stands for a relative pronoun. 

We evaluate bias toward [ANIMAL] by observing the change of predicted probability of words at the [MASK] token by replacing [PRONOUN] and [REL-PRONOUN].
We use the following combinations of [PRONOUN] and [REL-PRONOUN]:
 

\vspace{-.5\baselineskip}
\begin{itemize}
  \setlength{\parskip}{.05cm} 
  \setlength{\itemsep}{0cm} 
    \item human-describing sentences (hereinafter referred to as ``human sentences'')
    \begin{itemize}
        \item \textit{She} is a [ANIMAL] \textit{who} is [MASK].
        \item \textit{He} is a [ANIMAL] \textit{who} is [MASK].
    \end{itemize}
    \item object-describing sentences (hereinafter referred to as ``object sentences'')
    \begin{itemize}
        \item \textit{This} is a [ANIMAL] \textit{which} is [MASK].
        \item \textit{That} is a [ANIMAL] \textit{which} is [MASK].
        \item \textit{It} is a [ANIMAL] \textit{which} is [MASK].
        \item \textit{This} is a [ANIMAL] \textit{that} is [MASK].
        \item \textit{That} is a [ANIMAL] \textit{that} is [MASK].
        \item \textit{It} is a [ANIMAL] \textit{that} is [MASK].
    \end{itemize}
\end{itemize}
\vspace{-.5\baselineskip}
In human sentences, we use ``she'', ``he'', and ``who'', which generally refer to humans. In object sentences, we use ``this'', ``that'', ``it'', and ``which'', which are generally used for nonhumans. 
Since pronouns in object sentences are only in the third person equivalently, only the third person pronouns ``she'' and ``he'' are used in human sentences.

Our hypothesis here is that the characteristics of the words that are filled in ``[MASK]'' will change among animals that are often referred to in the speciesist language and others that are not.
For example, not only humans, but also dogs and cats could be referred to by the non-speciesist language, while ``farm animals'' (e.g. cow and pig) would be addressed by the speciesist language.

\subsubsection{Bias Evaluation by Word Probability Differences}
\label{subsec:bias-evaluation-probdiff}

We evaluate the bias against animal names using words with a large change rate of average predicted probability between human and object sentences.
It is done by averaging predicted probability of the word filled into the [MASK] token in the template sentences. 
We also investigate the relationship between animals by clustering them using the agreement rate of words with large probability changes.
We perform this experiment as follows:



\begin{enumerate}
  \setlength{\parskip}{.05cm} 
  \setlength{\itemsep}{0cm} 
    \item Calculating \reviewera{$p^{w_i}_{mean_o}(\rm name)$ and $p^{w_i}_{mean_h}(\rm name)$, which represent predicted probabilities of words corresponding to [MASK] tokens in each sentence, averaged over object ($p^{w_i}_{mean_o}$) and human ($p^{w_i}_{mean_o}$) sentences}, respectively, where $\rm name$ is an animal name and $w_i$ is a token in vocabulary $V$ of the MLM (i.e. $w_i \in V$)
    \item Calculating how much this probability changes by \modified{$\log \frac{p^{w_i}_{mean_o}(\rm name)}{p^{w_i}_{mean_h}(\rm name)}$}
    \item Ignoring words $w_i$ if (a) both \modified{$p^{w_i}_{mean_{o}}(\rm name)$ and  $p^{w_i}_{mean_{o}}(\rm name) < \frac{1}{|V|}$}, or \reviewera{(b) insignificant words at the center of the distribution of $\log \frac{p^{w_i}_{mean_o}(\rm name)}{p^{w_i}_{mean_h}(\rm name)}$ for each MLM}
    
    \item Calculating Token-Match-Rate (TMR) among animal names
    \item Clustering all animals based on TMR with UPGMA algorithm \citep{michener1957quantitative}.
\end{enumerate}
In step 1, we calculate $p^{w_i}_{mean_{o,h}}$ as follows:
\begin{equation}
\reviewera{p^{w_i}_{mean}({\rm name}) = \frac{1}{|T|}\sum_{s\in T}^{|T|} p_{\rm MLM}(w_i | s(\rm name))}
\end{equation}
where $T$ is the set of object or human template sentences described above, $s(\rm name)$ is a template sentence filled with an animal name. 
\reviewera{In step 3, in order to ignore insignificant words, first we calculate z-score of all words $w_i$ of $\log \frac{p^{w_i}_{mean_o}(\rm name)}{p^{w_i}_{mean_h}(\rm name)}$, and then we ignore words with $|\text{z-score}|$ lower than a threshold\footnote{\modified{Here this threshold value was experimentally set to 1.96.}}. \reviewera{Here z-score $= \frac{x-\mu}{\sigma}$, $x$ is $\log \frac{p^{w_i}_{mean_o}(\rm name)}{p^{w_i}_{mean_h}(\rm name)}$, $\mu$ is the average, $\sigma$ is the standard deviation.} 
}
In step 4,
where $S^{(i)}$ and $S^{(j)}$ are the obtained sets of words for the $i,j$-th animal names after step 3, we calculate TMR$(i,j)$ between both sets \citep[cf.][]{webster2020measuring-pretrained-models,lauscher2021sustainable-module-debiasig}:
\begin{equation}
\label{eq:token-match-rate}
\text{TMR}(i,j) = \frac{|S^{(i)} \cap S^{(j)}|}{\min(|S^{(i)}|,|S^{(j)}|)}
\end{equation}
In step 5, we cluster animal names by using $1-\text{TMR}(i,j)$ as distance between $i,j$-th names.

\subsubsection{Bias Evaluation by Sentiment Analysis}
\label{subsec:bias-eval-sent-analysis}
In this experiment, we use VADER \citep{hutto2014vader} for evaluating the sentiment of all words which we obtain from the experiment described in \secref{subsec:bias-evaluation-probdiff}.
This approach does not take into account context when evaluating sentiment of the words, but we decided to analyze the sentiment of the words themselves, considering the possibility of (non-)speciesist bias in the animal names.

Our hypothesis is that when animals are regarded as objects, they are treated negatively, and therefore more negative words will appear under MASKs in object sentences.

\subsection{Corpus-based Experiment}
\label{subsec:corpus-experiment}

In this section, we explain how the bias is measured in the corpus-based evaluation method.
The corpus used in this paper is Books3 \citep[see also \citep{gao2020pile-800GB}]{Books3} which totals about 100GB of text and is built only from published books. 
Thus, it is unlikely to overlap with BookCorpus, which contains unpublished books used for the pre-training of MLMs.

To experiment with corpus-based method, we extract object and human sentences from a given corpus.
For the purpose of this research, we extract all corpus sentences that contain relative pronouns referring to animals. We use five relative pronouns: ``that'', ``which'', ``who'', ``whose'' and ``whom''.
Our assumption is that these relative pronouns can be used to determine whether (non)human animals are treated as objects or humans in the given sentence.

CoreNLP \citep{manning-etal-2014-stanford-corenlp} is used to extract sentences containing relative pronouns which refer to an animal name.
If the speciesist bias exists in CoreNLP, then there may be a difference in referring precision between human and object sentences. 
Therefore, we asked a native speaker of English to check whether relative pronouns are correctly referred to an animal name in ten sentences (for each pronoun) randomly extracted from Book3.
As a result, one sentence containing ``who'', and two with ``whom'' have been marked as incorrect, and all remaining 47 sentences have been judged as having correct references.
This suggests that the precision of the parser for the task is relatively high.

For the corpus-based bias evaluation, we replace relative pronouns referring to animal names with [MASK] tokens in extracted sentences.
Then, we use MLMs to calculate probabilities of relative pronouns at the [MASK] token.
We compare the probabilities for both sets and evaluate the bias as follows:
\begin{equation}
\label{eq:bias_relative_pred}
bias =  \frac{1}{|H|} \sum_{s_i \in H}^{|H|} \mathbbm{1}[p_{object|s_i} > p_{human|s_i}] -\frac{1}{|O|} \sum_{s_j \in O}^{|O|} \mathbbm{1}[p_{human|s_j} > p_{object|s_j}] 
\end{equation}
where $H$ and $O$ are the sets of human and object sentences extracted from Books3, and $s_{i,j}$ is a given sentence.
$\mathbbm{1}[\cdot]$ returns 1 if its condition is true and 0 otherwise.
$p_{object|s_i}$ and $p_{human|s_i}$ are represented as follows:
\vspace{-.2\baselineskip}
\begin{align}
\begin{split}
&p_{object|s_i} = \max(p_{that|s_i}, p_{which|s_i}) \nonumber\\ 
&p_{human|s_i} = \max(p_{who|s_i}, p_{whose|s_i}, p_{whom|s_i}) 
\end{split}
\end{align}
Variables $p_{that|s_i},$ $p_{which|s_i},$ $p_{who|s_i},$ $p_{whose|s_i},$ and $p_{whom|s_i}$ are the probabilities of each relative pronoun substituting [MASK] in a given sentence.
If the value of the first term in the Equation \ref{eq:bias_relative_pred} is closer to 1, MLMs incorrectly predict higher probability of ``which'' or ``that'', and if the second term approaches 1, MLMs incorrectly predict higher probability of ``who'', ``whose'' or ``whom''.
 In other words, when the bias is close to 1, models tend to regard animals as objects; and if it is close to -1, they tend to treat them as humans.

 To investigate the relationship between the bias represented in Equation \ref{eq:bias_relative_pred} and the frequency bias in the corpora, we also calculate the correlation between the bias and the frequency of object-related pronouns (``that'' and ``which'') referring to each animal name in Wikipedia and BookCorpus.




\begin{figure}[p]
    \centering
    \begin{tabular}{c}
        \begin{minipage}{0.96\linewidth}
         \centering
         \includegraphics[trim=130 150 120 10, clip,width=\linewidth]{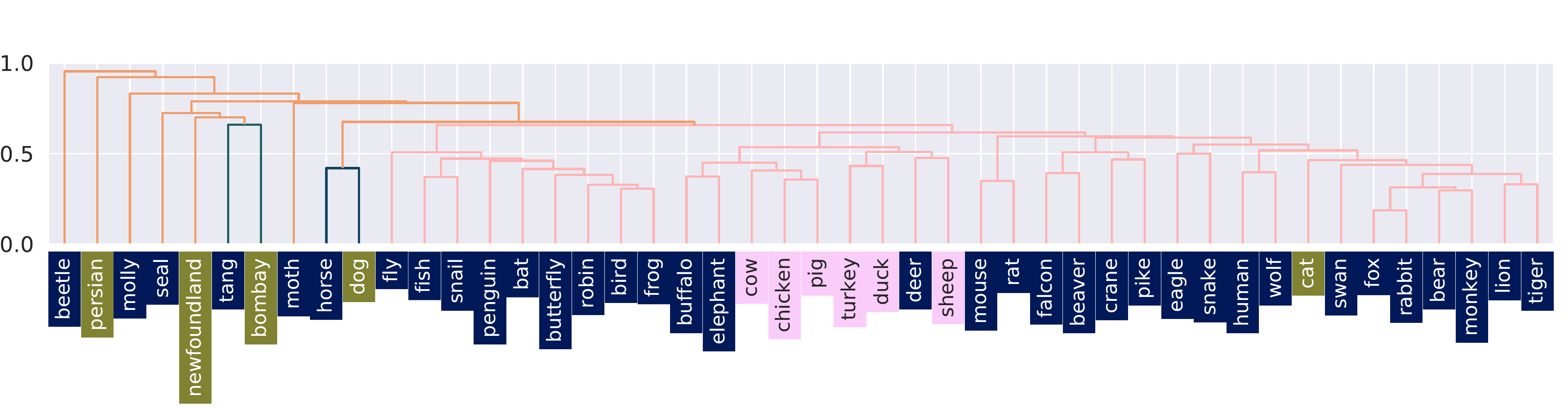}
         \subcaption{BERT}
         \label{fig:dendrogram-bert}
        \end{minipage} \\  
        \begin{minipage}{0.96\linewidth}
         \centering
         \includegraphics[trim=130 150 120 10, clip,width=\linewidth]{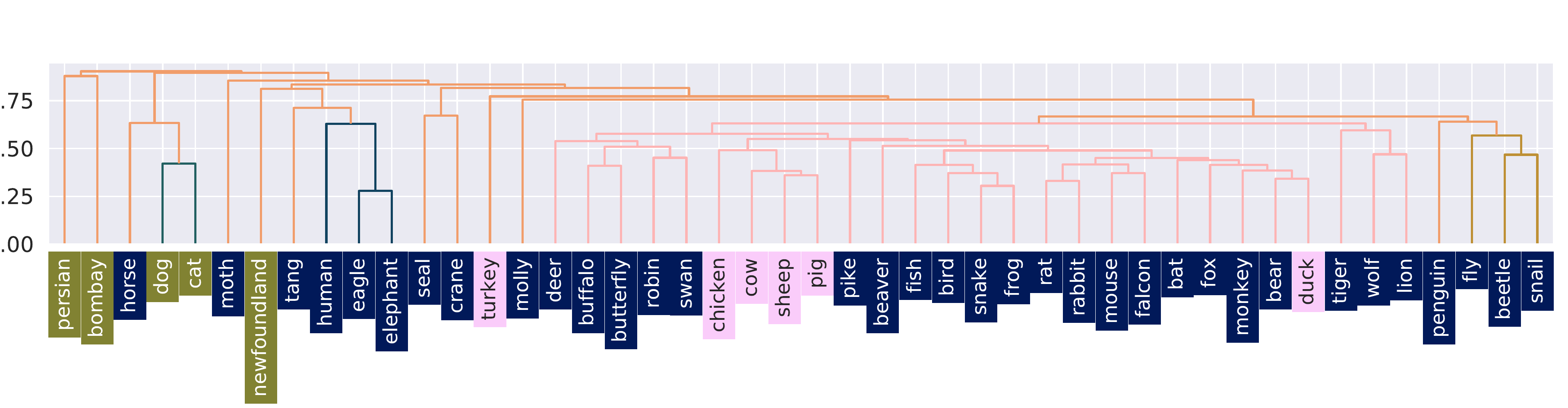}
         \subcaption{RoBERTa}
         \label{fig:dendrogram-roberta}
        \end{minipage} \\  
        \begin{minipage}{0.96\linewidth}
         \centering
         \includegraphics[trim=130 150 120 10, clip,width=\linewidth]{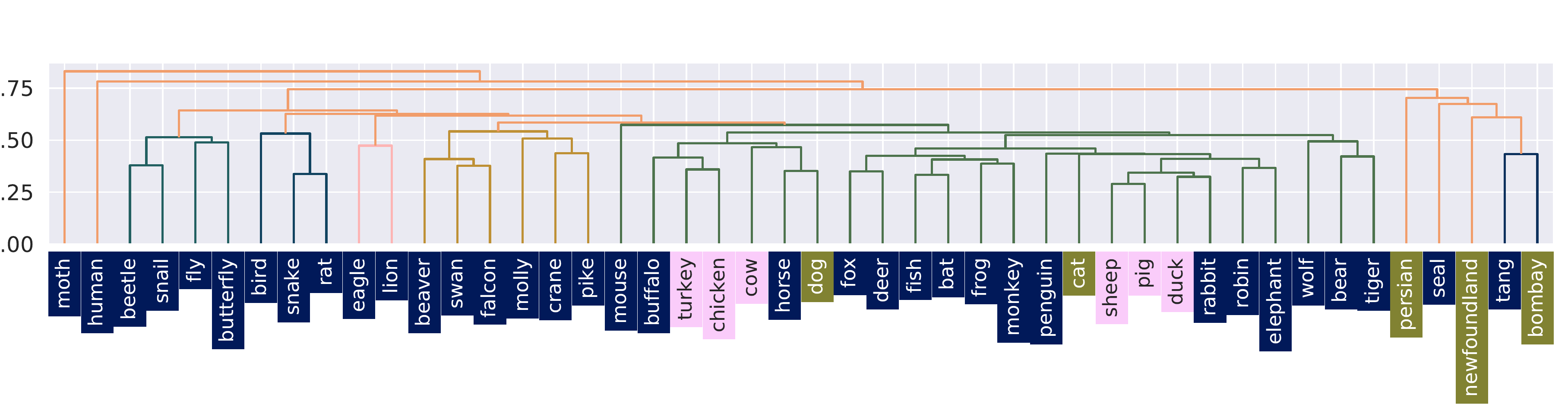}
         \subcaption{DistilBERT}
         \label{fig:dendrogram-distil}
        \end{minipage} \\
        
        \begin{minipage}{0.96\linewidth}
         \centering
         \includegraphics[trim=130 150 120 10, clip,width=\linewidth]{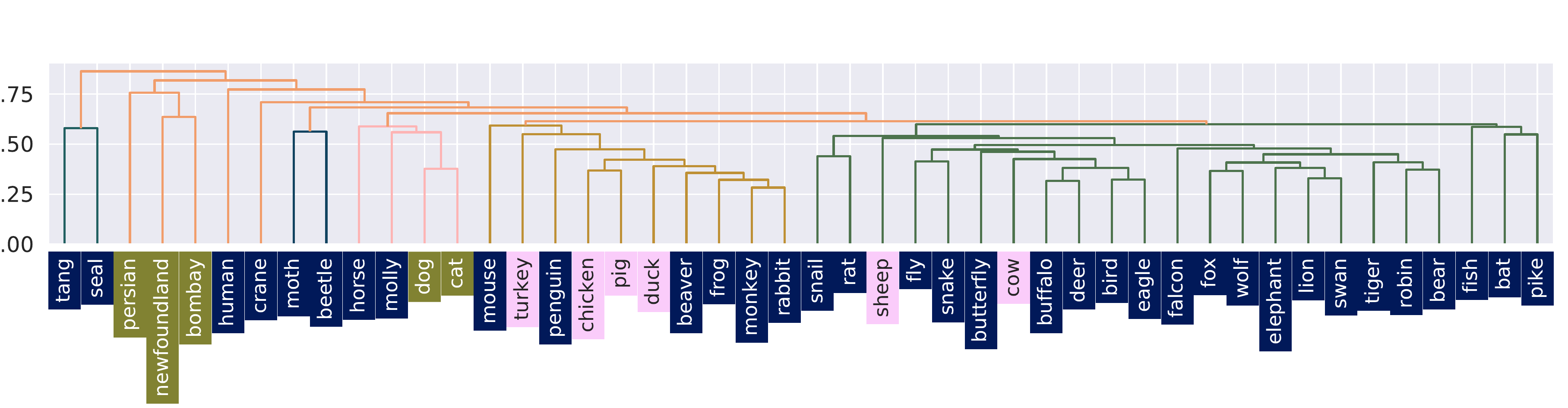}
         \subcaption{ALBERT}
         \label{fig:dendrogram-albert}
        \end{minipage}
    \end{tabular}
    \caption{Results of hierarchical clustering based on the agreement rate of words whose predicted probability of filling the [MASK] token changed significantly between template sentences. \reviewera{The vertical axis represents $1-$TMR by Equation \ref{eq:token-match-rate}. TMR indicates similarity among animal names.} Each leaf is colored using SciPy library \citep{2020SciPy-NMeth}, with the default color threshold. \reviewerb{ \colorboxself{\thinpink} refers to ``farm'' animals, \colorboxself{\dirtyyellow} indicates nonhuman companions and \colorboxself{\blueblack}  stands for the remaining species.}}
    \label{fig:dendrogram}
\end{figure}

\section{Experimental Results}
\label{sec:results}

In this section, we first report the results of the template-based experiments (\secref{subsec:result-template-based}), and then the results of the corpus-based experiment (\secref{subsec:result-corpus-based}).

\subsection{Template-based Evaluation}
\label{subsec:result-template-based}
\subsubsection{Probability Differences}

The experimental results of probability differences between human and object sentences are presented in \figref{fig:dendrogram}, and Figures \ref{fig:heatmap-bertl}, \ref{fig:hearmap-roberta}, \ref{fig:hearmap-distil}, \ref{fig:hearmap-albert} in Appendix \ref{sec:appendix}.

From \figref{fig:dendrogram} it can be observed that the names of animals colored with the same color belong to roughly the same clusters. Especially in the results of BERT and RoBERTa, the names of animals who are often kept at farms were clustered closely in most cases (see Figures \ref{fig:dendrogram-bert} and \ref{fig:dendrogram-roberta}).
In the cases of DistilBERT and ALBERT, the animal names with the same color were not grouped together, but some belonged to the same cluster, indicating that they were not completely disjointed.

In Tables \ref{tab:high_change_rate_BERT} and \ref{tab:high_change_rate_RoBERTa},  we show sets of top five words with the largest probability change for each animal. For these tables we chose the five most frequent animal names in Wikipedia, and added the most popular animals living in farms and at homes, as they are one of the focal interest of our investigation: ``cat'', ``dog'', ``chicken'' and ``pig''. In these tables, we show the results for BERT and RoBERTa, while the results of the remaining models are given in Appendix \ref{sec:appendix}.

For ``chicken'', ``pig'' and ``turkey'', words with high probability change in object sentences included ``slaughtered'', ``reproduced'', ``ripe'' (see \tabref{tab:high_change_rate_BERT}), also ``dried'' and ``harvested'' (see \tabref{tab:high_change_rate_RoBERTa}).
Also, in BERT, ``f**k''-rooted words were associated with many animals.
On the other hand, in human sentences, associated words express personality and gender-related attributes, such as ``clumsy'' or ``bisexual''.
There are also many words that represent personality traits that can be interpreted as negative, for example ``sarcastic''.
On the other hand, ``human'' did not exhibit many such characteristics, so the bias observed in the case of nonhuman animals was not caused by a bias in the template sentence itself, but a bias among the sentence and the name.

\subsubsection{Sentiment Analysis}
\label{subsubsec:result-sentiment-analysis}

\begin{figure*}
    \centering
    \includegraphics[trim=35 13 10 10,clip,width=0.99\linewidth]{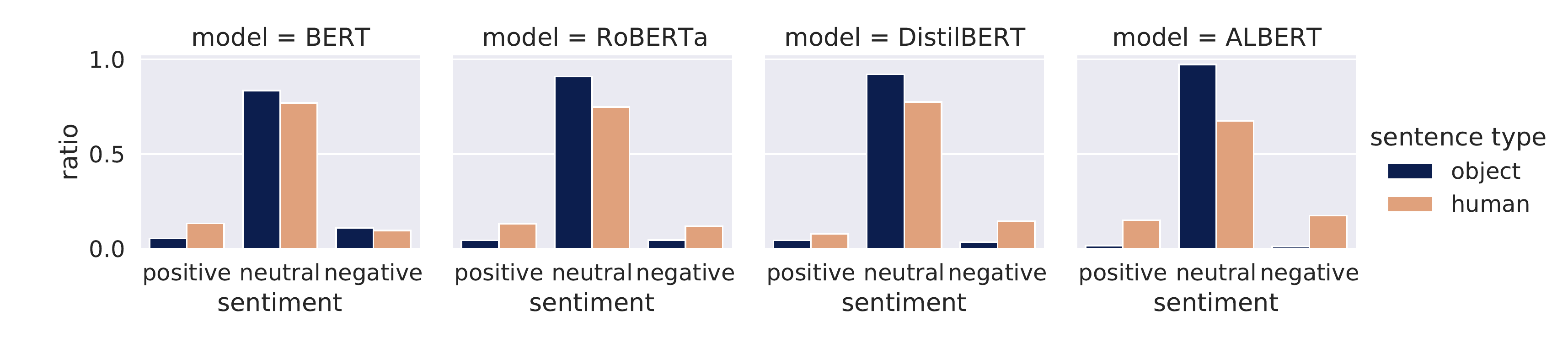}
    \caption{Results of sentiment analysis for each language model. Vertical axis shows the ratio of words assigned to a certain sentiment. For each sentiment, the darker bars indicate the percentage of words that have a higher mean probability in the object sentences, and the light-colored ones show the ratio of words that have a higher mean probability in the human sentences.}
    \label{fig:sentiment-bias}
\end{figure*}

Next, we report the results of the sentiment analysis performed on obtained words in the experiment described in \ref{subsec:bias-eval-sent-analysis} (see \figref{fig:sentiment-bias}).
The vertical axis of the figure shows the percentage of the number of words assigned to each sentiment.
The horizontal one shows the sentiment and the names of the models.

We found that VADER assigned 0 (i.e. neutral sentiment) to the majority of the words, and
that object sentences contained more neutral words than human sentences in all models.
Contrary to our hypothesis, the ratio of negative words was found to be larger in human sentences for all three models except BERT.
Within each model, the distribution of assigned sentiment was generally the same.



\begin{table}[t]
    \centering
    \caption{Sets of five predicted words with the highest change rate in BERT. Possibly harmful biased words are shown in \textbf{bold} font.
    For example, it is clear that ``f**ked'' and ``slaughtered'' can be harmful to non-human animals. Also, ``ripe'' and ``beaf'' can be harmful to some nonhuman animals, assuming that they are not food. \modified{ \colorboxself{\thinpink} refers to ``farm'' animals, \colorboxself{\dirtyyellow} indicates nonhuman companions and \colorboxself{\blueblack}  stands for the remaining species.}
    }
    \footnotesize
    \scalebox{1}{
    \begin{tabularx}{1\linewidth}{|l|X|X|}
    \hline
        Animal name& Words with high probability change in \textit{object} sentences & Words with high probability change in \textit{human} sentences \\ \hline\hline
         \colorboxself{\dirtyyellow} cat& \textbf{f**ked}, \textbf{f**king}, \textbf{reproduced}, violated, \textbf{ripe} & \textbf{sarcastic}, \textbf{mute}, Ninja, \textbf{clumsy}, unnamed \\ \hline
         \colorboxself{\dirtyyellow} dog& \textbf{f**ked}, \textbf{f**king}, struck, violated, committed & \textbf{sarcastic}, Ninja, \textbf{mute}, bisexual, unnamed \\ \hline
         \colorboxself{\thinpink} chicken& \textbf{slaughtered}, \textbf{f**ked}, \textbf{stamped}, \textbf{reproduced}, \textbf{ripe} & \textbf{clumsy}, \textbf{mute}, \textbf{sarcastic}, psychic, superhero \\ \hline
         \colorboxself{\thinpink} pig& \textbf{f**ked}, \textbf{stamped}, \textbf{slaughtered}, \textbf{reproduced}, \textbf{sin} & \textbf{clumsy}, \textbf{sarcastic}, \textbf{mute}, cheerful, blonde \\ \hline
         \colorboxself{\thinpink} turkey& \textbf{stamped}, \textbf{slaughtered}, \textbf{beef}, \textbf{ripe}, viable & \textbf{mute}, \textbf{clumsy}, psychic, \textbf{sarcastic}, \textbf{deaf} \\ \hline
         \colorboxself{\blueblack} fish& endemic, predatory, widespread, perennial, barred & heroine, \textbf{sarcastic}, Cinderella, princess, cheerful \\ \hline
          \colorboxself{\blueblack} fox& \textbf{f**ked}, happening, waking, calling, ours & \textbf{mute}, \textbf{sarcastic}, blonde, bisexual, \textbf{clumsy} \\ \hline
          \colorboxself{\blueblack} horse& \textbf{f**ked}, \textbf{sin}, violated, \textbf{stamped}, \textbf{ripe} & unnamed, pink, \textbf{sarcastic}, blonde, Ariel \\ \hline
          \colorboxself{\blueblack} human& ourselves, worth, ours, yours, our & bisexual, Ninja, \textbf{sarcastic}, blonde, lesbian \\ \hline
    \end{tabularx}
    }
    \label{tab:high_change_rate_BERT}
\end{table}

\begin{table}[t]
    \centering
    \caption{Sets of five predicted words with the highest change rate in RoBERTa. Possibly harmful biased words are shown in \textbf{bold} font. \modified{ \colorboxself{\thinpink} refers to ``farm'' animals, \colorboxself{\dirtyyellow} indicates nonhuman companions and \colorboxself{\blueblack}  stands for the remaining species.}}
    \footnotesize
    \scalebox{1}{
    \begin{tabularx}{1.0\linewidth}{|l|X|X|}
    \hline
        Animal name & Words with high probability change in \textit{object} sentences & Words with high probability change in \textit{human} sentences \\ \hline\hline
         \colorboxself{\dirtyyellow} cat& terrestrial,  armoured,  netted,  scaled,  predatory & foster,  \textbf{deaf},  Transgender,  Blind,  Polish \\ \hline
         \colorboxself{\dirtyyellow} dog& terrestrial,  itself,  predatory,  defined,  armoured &  \textbf{deaf},  transsexual,  foster,  Homeless,  lesbian \\ \hline
         \colorboxself{\thinpink} chicken& \textbf{dried},  freshwater,  semen,  \textbf{polled},  \textbf{harvested} & optimistic,  \textbf{sarcastic},  romantic,  pessimistic,  Psychic \\ \hline
         \colorboxself{\thinpink} pig&  \textbf{polled},  \textbf{dried},  \textbf{harvested},  yielded,  \textbf{peeled} &  romantic,  \textbf{selfish},  optimistic,  jealous,  \textbf{arrogant} \\ \hline
         \colorboxself{\thinpink} turkey&  \textbf{dried},  \textbf{processed},  ground,  \textbf{slaughtered},  cached, &  \textbf{deaf},  listening,  jealous,  optimistic,  psychic \\ \hline
         \colorboxself{\blueblack} fish& freshwater,  reef,  widespread,  \textbf{polled},  aggregate & \textbf{swearing},  jealous,  witty,  superhuman,  sixteen \\ \hline
          \colorboxself{\blueblack} fox&  \textbf{polled},  invasive,  Madagascar,  pictured,  extant &  pessimistic,  \textbf{sarcastic},  \textbf{mercenary},  romantic,  compassionate \\ \hline
          \colorboxself{\blueblack} horse&  clicking,  enough,  beat,  it,  right &  Transgender,  lesbian,  \textbf{deaf},  transgender,  transsexual \\ \hline
          \colorboxself{\blueblack} human&  extant,  extinct,  ours,  yours,  \textbf{edible} &  bartender,  nineteen,  seventeen,  sixteen,  eighteen \\ \hline
    \end{tabularx}
    }
    \label{tab:high_change_rate_RoBERTa}
\end{table}


\begin{table}[tb]
    \centering
    \caption{Frequency of relative pronouns referring to animal names in each corpus (references determined by CoreNLP). The number in parentheses is the total number minus the number indicating ``\reviewera{human}''.}
    \small
    \scalebox{1}{
    \begin{tabular}{|l|p{1.2cm}|p{1cm}|p{1cm}|p{0.9cm}|p{0.9cm}|}
    \hline
        Corpus & that & which & who & whose & whom \\ \hline\hline
        Books3 & 104,244 (103,361) & 28,552 (28,231) & 44,607 (39,593) & 4,115 (4,012) & 2,006 (1,690) \\ \hline
        Books-Corpus & 5,111 (4,949) & 1,470 (1,419) & 3,925 (2,988) & 183 (171) & 66 (50) \\ \hline
        Wikipedia (EN) & 9,341 (9,265) & 6,642 (6,586) & 7,182 (6,648) & 411 (396) & 289 (274)\\ \hline
    \end{tabular}
    }
    \label{tab:total-num-relativepro}
\end{table}

\subsection{Corpus-based Evaluation}
\label{subsec:result-corpus-based}

Here, we present the results of the corpus-based experiment. First, we look at the sentences extracted from the corpora.
In \tabref{tab:total-num-relativepro} we show the total numbers of relative pronouns referring to animal names in each corpus. 
The numbers in brackets indicate the total number minus the number of relative pronouns referring to ``human''. 
The total numbers for each animal are shown in Figures \ref{fig:animal_rel_ref_freq_in_wiki}, \ref{fig:animal_rel_ref_freq_in_bookscorpus} and \ref{fig:animal_rel_ref_freq_in_books3}.
Comparing the total numbers of ``that'' and ``which'' with the total numbers of ``who'', ``whose'' and ``whom'', we found that the former group is about twice more common. This indicates that the corpus as a whole tends to treat nonhuman animals as objects.
In addition, contrary to our assumption, the number of relative pronouns such as ``who'' that refers to ``dogs'' and ``cats'' in all corpora is almost the same as the total number of ``that'' and ``which'' (see Figures \ref{fig:animal_rel_ref_freq_in_wiki} and \ref{fig:animal_rel_ref_freq_in_bookscorpus}).

\begin{figure}[tb]
    \centering
    \begin{tabular}{cc}
        \begin{minipage}{0.48\linewidth}
         \centering
         \includegraphics[trim=95 25 95 20, clip,width=\linewidth]{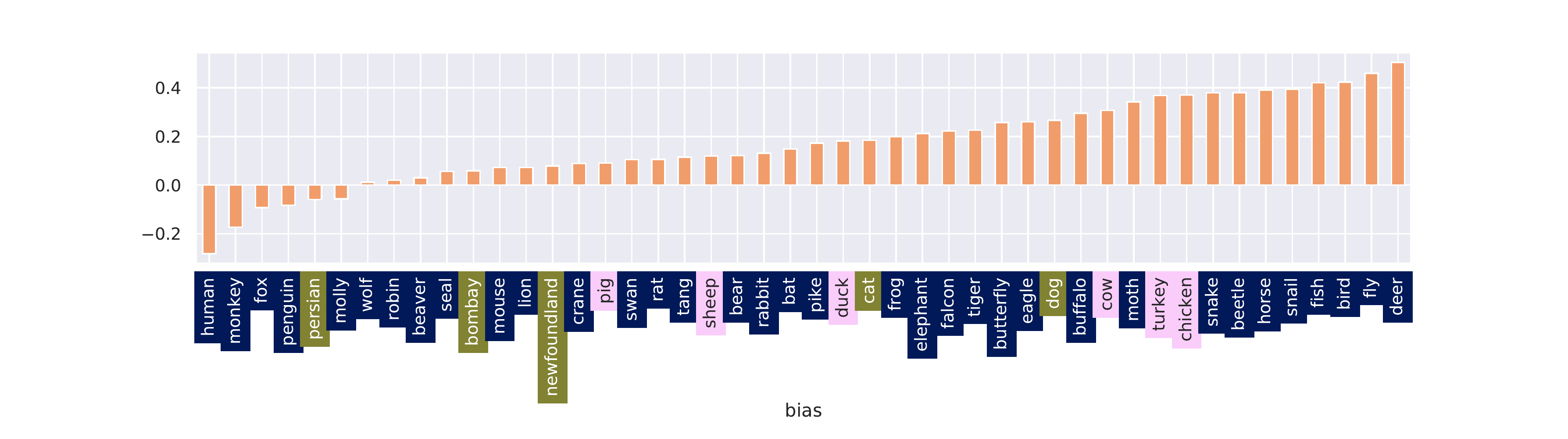}
         \subcaption{BERT}
         \label{fig:bias-bert}
        \end{minipage} &  
        \begin{minipage}{0.48\linewidth}
         \centering
         \includegraphics[trim=95 25 95 20, clip,width=\linewidth]{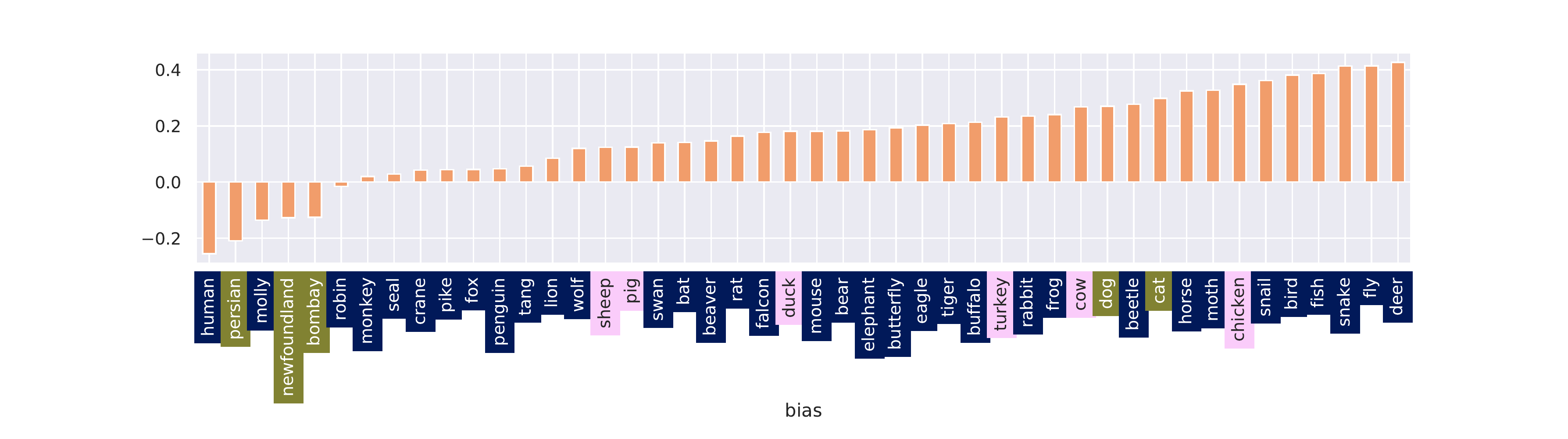}
         \subcaption{RoBERTa}
         \label{fig:bias-roberta}
        \end{minipage}\\
        \begin{minipage}{0.48\linewidth}
         \centering
         \includegraphics[trim=95 25 95 20, clip,width=\linewidth]{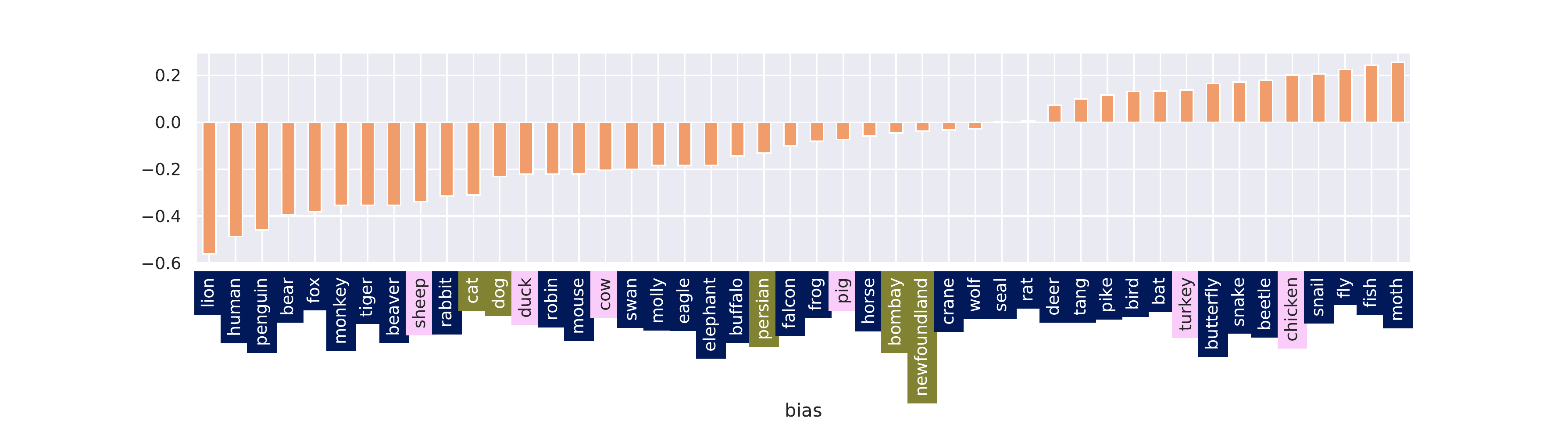}
         \subcaption{DistilBERT}
         \label{fig:bias-distil}
        \end{minipage} 
          &  
        \begin{minipage}{0.48\linewidth}
         \centering
         \includegraphics[trim=95 25 95 20, clip,width=\linewidth]{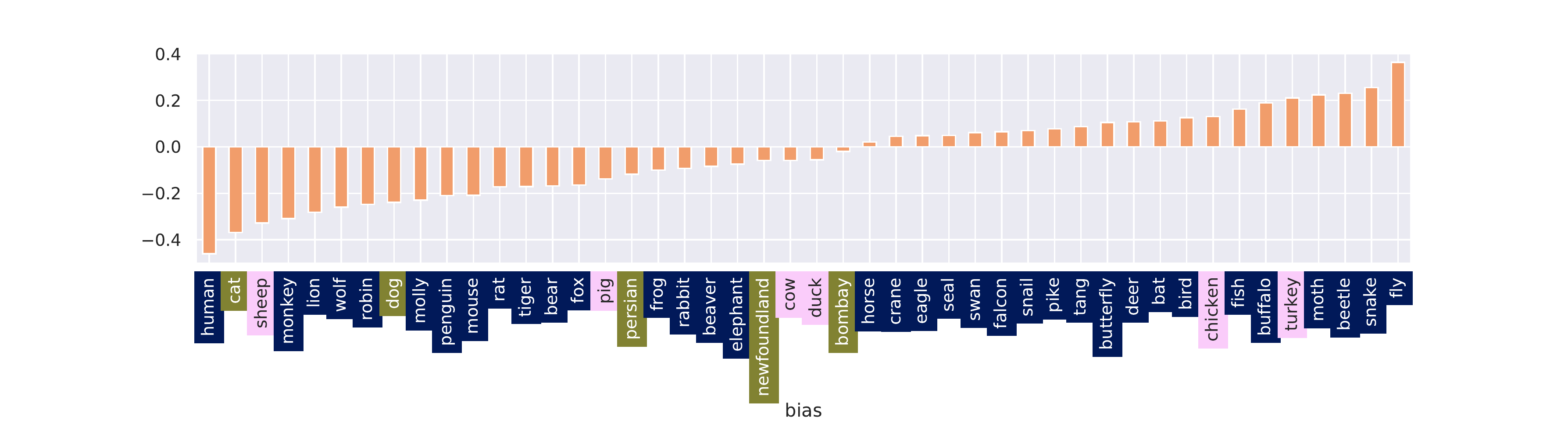}
         \subcaption{ALBERT}
         \label{fig:bias-albert}
        \end{minipage}
    \end{tabular}
    \caption{Results of the corpus-based bias analysis, sorted by the magnitude of the bias represented by Equation \ref{eq:bias_relative_pred}. Vertical axis shows the magnitude of the bias, where positive values indicate that MLMs incorrectly insert ``that'' or ``which'', and negative values indicate that MLMs incorrectly insert ``who'', ``whose'', or ``whom'' with higher probability. The horizontal one shows the animal names. \modified{Larger} versions of the graphs are given in Appendix \ref{sec:appendix}. \reviewerb{ \colorboxself{\thinpink} refers to ``farm'' animals, \colorboxself{\dirtyyellow} indicates nonhuman companions and \colorboxself{\blueblack}  stands for the remaining species.}
    }
    \label{fig:biasinbook3}
\end{figure}

Next, we examine the results of analyzing the bias of MLMs using sentences collected from the Books3 corpus (see \figref{fig:biasinbook3}).
The vertical axis of each graph represents the degree of bias, and the horizontal one represents the animal names. A positive bias indicates a high probability of incorrectly entering ``that'' or ``which'' (i.e., having a speciesist bias), while a negative bias indicates a high probability of incorrectly filling ``who'', \reviewera{``whose'', or ``whom''} (i.e., having a non-speciesist bias).

All of the models exhibited a negative bias against ``human'', and a positive bias against ``chicken'' and ``turkey''. These results are in line with our expectations. 
However, contrary to our predictions, the bias for ``dog'' and ``cat'' in BERT and RoBERTa is positive, indicating that they tend to be treated as objects. 
On the other hand, DistilBERT and ALBERT were found to include more negative bias, i.e. non-speciesist tendency, compared to BERT and RoBERTa.
\tabref{tab:correlation-bias-freq} shows the correlation between these biases and the ratio of the frequency of object-related pronouns in the corpora.
The correlation was above 0.7 for MLMs other than RoBERTa, and above 0.5 for RoBERTa, which indicates that the ratio of relative pronouns in the corpus explains the bias of MLMs to some extent.
We think that the low value for RoBERTa is due to the fact that RoBERTa has been pre-trained on other corpora.

\begin{table}[t]
    \centering
    \caption{Pearson correlation coefficient ($r$) between the bias represented in Equation \ref{eq:bias_relative_pred} and frequency of object-related pronouns in Wikipedia and BookCorpus.}
    \small
    \scalebox{1}{
    \begin{tabular}{|c|c|c|c|c|}
    \hline
         & BERT & RoBERTa & DistilBERT & ALBERT \\ \hline
        $r$ & 0.77  & 0.55  & 0.81  & 0.74  \\ \hline
    \end{tabular}
    }
    \label{tab:correlation-bias-freq}
\end{table}

\section{Discussion}
\label{sec:discussion}
\subsection{Template-based Approach}

The results of the animal names clustering in BERT and RoBERTa partially support the hypothesis of this experiment that MLMs vary animal-related words between \reviewera{object} and human sentences depending on the animal name.
On the other hand, DistilBERT and ALBERT performed clustering slightly different from our expectation, which may be due to the lower performance of mask predictions caused by the smaller model size.

\begin{table}[t]
    \centering
    \caption{\reviewera{Pseudo-perplexity (PPPL) on pseudo-log-likelihood score (PLL) \citep{salazar-etal-2020-masked-score} on 1,000 sentences, which are randomly sampled from collected sentences in \secref{subsec:corpus-experiment}. Lower value indicates better performance of mask predictions.}}
    \small
    \scalebox{1}{
    \begin{tabular}{|c|c|c|c|c|}
    \hline
         & BERT & RoBERTa & DistilBERT & ALBERT \\ \hline
        PPPL & 1.062  & 1.105  & 1.109  & 1.113  \\ \hline
    \end{tabular}
    }
    \label{tab:mlmscore-books3}
\end{table}
\reviewera{
To measure performance of mask predictions of each model, we calculate pseudo-perplexity (PPPL) on pseudo-log-likelihood score (PLL) \citep{salazar-etal-2020-masked-score} of each model (see \tabref{tab:mlmscore-books3}) on 1,000 sampled sentences from collected text for performing Corpus-based experiment (\secref{subsec:corpus-experiment}).
BERT achieves the highest score, and other model's scores yield similar. These results may partially explain why BERT clustering mirrored our expectations. while RoBERTa clustering cannot be explained by the mask prediction performance.
}

As shown in Tables \ref{tab:high_change_rate_BERT} and \ref{tab:high_change_rate_RoBERTa}, when nonhuman animals are described by object sentences, they are linked with harmful words such as ``f**ked''. 
Furthermore, in the case of animals who live in farms to be utilized as flesh, meat-related words have been confirmed, for example ``slaughtered'' and ``harvested'' described as problematic in previous studies \citep[][see also \secref{subsec:speciesism-and-language}]{dunayer2001animal-equality,dunayer_2003_english_and_spe}.
These words are likely to be associated with speciesist language that objectifies animals.
In addition, contrary to the hypothesis of the present experiment, harmful words were also associated with human sentences. This suggests a speciesist bias in which MLMs treat nonhuman animals negatively, even if they are treated as humans.

In the experiments of sentiment analysis, 
it is important to note here that VADER itself may exhibit a speciesist bias.
For example, VADER considers ``killed'' to be a negative word, but recognizes ``slaughtered'' as a neutral word. This problem should be investigated further. 

\subsection{Corpus-based Approach}

Frequencies of human-related pronouns are lower than object-related pronouns in all corpora (see \tabref{tab:total-num-relativepro}).
There are at least two possible causes for this discrepancy: (1) there are fewer human-related relative pronouns that refer to nonhuman animals in the corpus than object-related ones, or (2) the recall of CoreNLP for human-related relative pronouns is low.
If (1) is correct, it suggests that people tend to treat nonhuman animals as objects. If (2) is correct, it suggests that there is a bias in CoreNLP which makes the parser unable to sufficiently capture human-related relational references to nonhuman animals. Either result could be indirectly harmful to nonhuman animals.

In our corpus bias evaluation experiments, we found that, contrary to our hypothesis, the models had a speciesist bias against ``dog'' and ``cat''. However, all models exhibited a non-speciesist bias for more specific kinds of dogs and cats such as \reviewera{``newfoundland''} and ``persian''. 
These results suggest that MLMs predicted ``that'' and ``which'' referring to ``dog'' and ``cat'' with high probability  because they are commonly used as general names and therefore do not represent specific individuals. 
The bias between general names and more specific names will also be a subject of our future work. 

\subsection{Limitations}
In all of these experiments, the speciesist bias is assessed using a difference between the object and human sentences. Therefore, it is not possible to evaluate biases that do not appear in that difference. Especially, we cannot evaluate words or concepts that are simply associated with animal names in this experiment. Since the purpose of this research is to investigate the relationship between animal names and speciesist language and its bias, it is necessary to evaluate other types of bias related to animal names in the future. 

In addition, the corpus used in the corpus-based experiment is Books3, a corpus consisting of published books, so it is possible that different biases will be detected in different types of corpora. However, since the magnitude of this bias is correlated with the frequency of occurrence in Wikipedia and BookCorpus, we think that the same results are likely to be confirmed when using other textual resources.

\section{\reviewerb{How to Mitigate Speciesist Bias?}}
\label{sec:mitigate-bias}
\reviewerb{Based on the experimental results of this paper (and \citet{hagendorff2022speciesist-bias}), we briefly discuss possible ways to mitigate speciesist bias. Here we consider, according to the classification of the \citet{shah-etal-2020-predictive-bias-nlp-framework}, label bias, selection bias, overamplification and semantic bias as causes of the speciesist bias. Note that the ideal data distribution is a non-speciesist distribution, i.e. the strength of the relationship between nonhuman animals and negative concepts or speciesist language is significantly weaker than the strength of the relationship between positive concepts or non-speciesist language.}

\reviewerb{First, regarding the \textit{label bias}, this bias emerges when there is discrepancy of the dependent variable (i.e. label) between source and the ideal distribution. Our experimental results are label-independent, so the speciesist bias found in this paper is not caused by the label bias. 
We did not analyze bias in downstream tasks such as sentiment analysis or text generation. 
However, it is possible to identify speciesism bias caused by label bias in the behavior of the NLP models used in downstream tasks.
For example, as mentioned in \secref{subsubsec:result-sentiment-analysis}, VADER (which is not a pre-trained language model) labels ``slaughter'' as neutral, meaning it contains a speciesist bias due to label bias. If a dataset for the sentiment analysis task contains a similar bias, then the behavior of models trained on that dataset may exhibit a speciesist bias.
One way to mitigate the label bias is to create non-speciesist annotation guidelines for downstream tasks directly or indirectly related to nonhuman animals.}


\reviewerb{Second, regarding the \textit{selection bias}, this bias occurs when there is a difference between source data distribution and ideal one.
Current English resources are considered to be speciesist. Therefore, at least two mitigation methods are possible.
The first method is to mitigate the bias of a corpus itself for example by using algorithms such as Counterfactual Data Augmentation (CDA)~\citep{zhao-etal-2018-gender-coreference, webster2020measuring-pretrained-models}. 
However, unlike in the cases of gender or race, because we do not know what attributes nonhuman animals correspond to, utilizing CDA would be problematic.
The second approach is to deliberately select a corpus that uses non-speciesist language. Here, news articles and web content that ethical vegans tend to contribute may be useful. \citet{devinney-etal-2020-semisupervised-topic-model-gender-bias} has created Queer corpus from news and web content relating to LGBTQ+ people and their issues and found that the topics discussed in Queer corpus differ from the main stream English corpus. Perhaps, topics in textual data discussing ethical vegan-related topics also differ from the main stream English corpus, and there are fewer speciesist language.}

\reviewerb{Third, we consider \textit{overamplification}, which occurs when, nevertheless the distribution of source data and ideal one are almost identical, a model amplifies small differences. The experimental results in this paper show that BERT, DistilBERT, and ALBERT associate different words with animal names, even though they were pre-trained on the same corpus. In particular, BERT associates more ``f**k''-rooted words, and DistilBERT and ALBERT assign the same word to many animal names. This phenomenon is unlikely to be caused by word distribution in the corpus, and is probably mainly due to overamplification. On the other hand, RoBERTa appears to assign a wide variety of words, and since corpora other than Wikipedia and BookCorpus were used for pre-training RoBERTa, it may be important to pre-train language models on many diverse corpora.}

\reviewerb{Forth type of bias which could be mitigated is \textit{semantic bias} which relates to ``unintended or undesirable associations and social stereotypes''~\citep{shah-etal-2020-predictive-bias-nlp-framework}. 
Many methods for mitigating and analyzing this bias require specific word lists~\citep[e.g.][]{liang-etal-2020-towards-debias-sentence-repr, kaneko-bollegala-2021-debiasing-pretrained-contect-embed}, but because some understanding of speciesism is needed to create them and many people are not even aware of the problem, recruting annotators would be difficult (also see the beginning of \secref{sec:bias-evaluation-method}). 
Therefore, bias mitigation methods such as Dropout~\citep{webster2020measuring-pretrained-models} and Self-debias~\citep{Schick2021-self-diagnosis-debias}, which do not require a specific word list, may be useful~\citep[cf.][]{meade-etal-2022-empirical-surve-effect-debias}.}

\reviewerb{Our experiments have also shown that biases differ among nonhuman animal species. Therefore, in addition to mitigating speciesist bias with humans, there is also a need to mitigate the bias among nonhuman animals. This may be difficult to do with methods described as above, therefore this problem should be addressed in the future.}
\section{Conclusion}
\label{sec:conclusion}

In this paper, we analyzed the speciesist bias against animals inherent in MLMs.
Our experimental results showed that such models strongly associate harmful words with many nonhuman animals. 
\reviewera{Especially for BERT and RoBERTa, these models clustered the animal names as we expected, and also associated the same harmful words (e.g. ``slaughtered'') with nonhuman animals who live in the farm.}
We also found that MLMs, especially BERT and RoBERTa, are biased to associate object-related pronouns (``that'' and  ``which'') with various nonhuman animals, and demonstrate that this bias is correlated with the frequency of these relative pronouns referring to each animal in the corpora.
Since this research is restricted to English language, it cannot be generalized to other languages.
Moreover, this paper does not address so-called \textit{intersectional bias}. 
For example, ``bitch'' means a female dog (intersectional attributes of gender and animal species), but it is also used as an insult toward women. 
In future, we plan to expand our research by utilizing findings in animal ethics regarding intersectional bias and discrimination between speciesist bias and other biases \citep{birke1995animals-and-women,adams1990sexual}.

\printcredits

\bibliographystyle{cas-model2-names}

\bibliography{mybib}

\appendix
\section{Appendix}
\label{sec:appendix}

\newpage
\begin{table}[h]
    \centering
    \caption{\reviewerb{Hyperparameter information regarding models using all experiments in this paper}}
    \begin{tabular}{|c|c|c|c|c|}
    \hline
       & BERT$_{\rm LARGE\text{-}cased}$ & RoBERTa$_{\rm LARGE}$ & DistilBERT$_{\rm base\text{-}cased}$ & ALBERT$_{\rm large\text{-}v2}$ \\ \hline\hline
        Parameters & 334M & 334M & 65M & 18M \\ \hline
        Layers & 24 & 24 & 6 & 24 \\ \hline
        Hidden & 1024 & 1024 & 768 & 1024 \\\hline
        Embedding & 1024 & 1024 & 768 & 128 \\\hline
        Other & - & - & distillation model of BERT & using parameter sharing \\\hline
    \end{tabular}
    \label{tab:model-details}
\end{table}

\begin{figure}[p]
    \centering
    \includegraphics[width=0.98\linewidth]{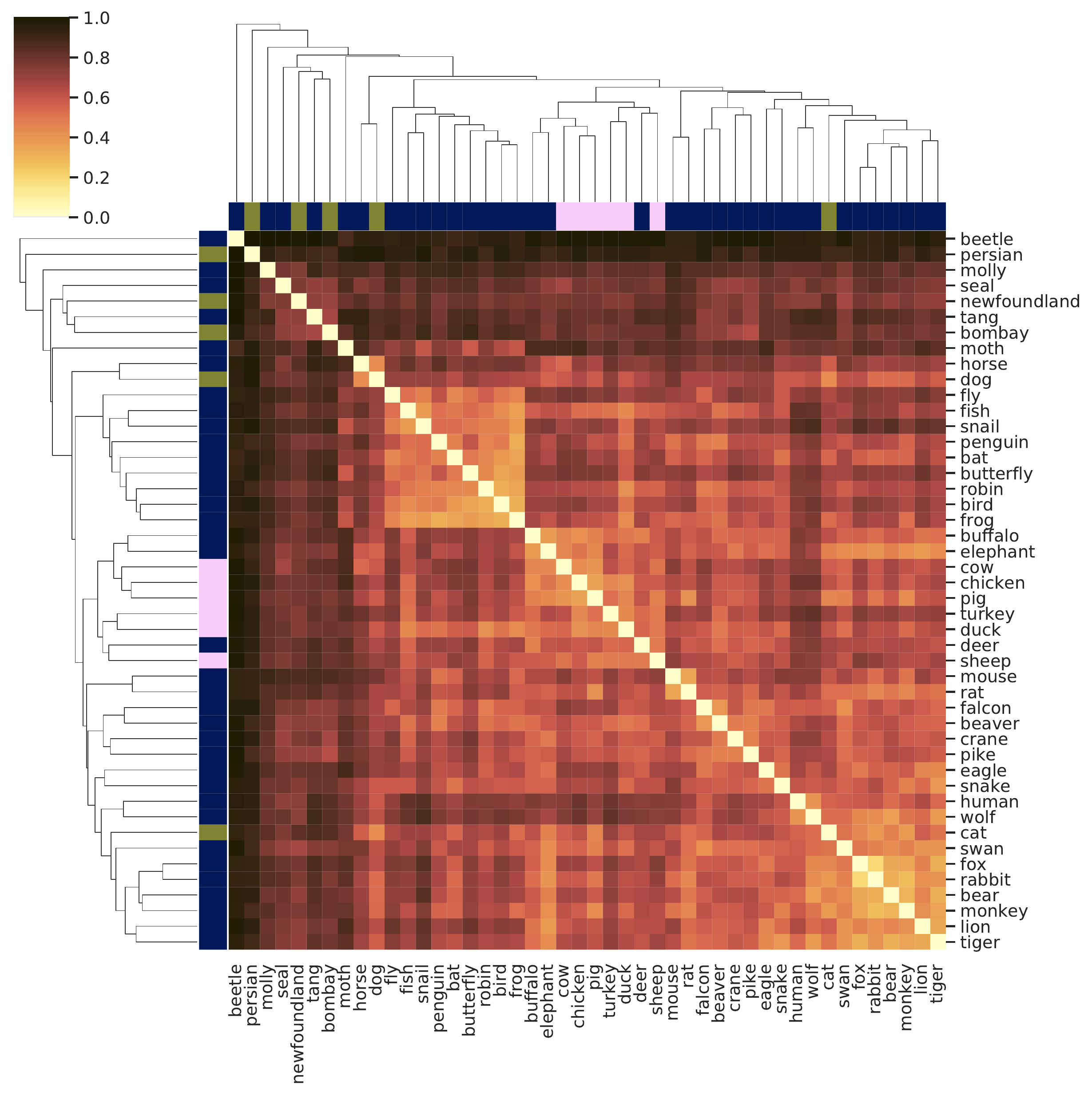}
    \caption{A heat map of the results of the template-based experiments, clustered by TMR with large probability changes in BERT: \colorboxself{\thinpink} refers to ``farm'' animals, \colorboxself{\dirtyyellow} indicates nonhuman companions and \colorboxself{\blueblack}  stands for the remaining species. \reviewera{The score represents the $1-$TMR. The lighter cells, the higher similarity among animals.}}
    \label{fig:heatmap-bertl}
\end{figure}

\begin{figure}[p]
    \centering
    \includegraphics[width=0.98\linewidth]{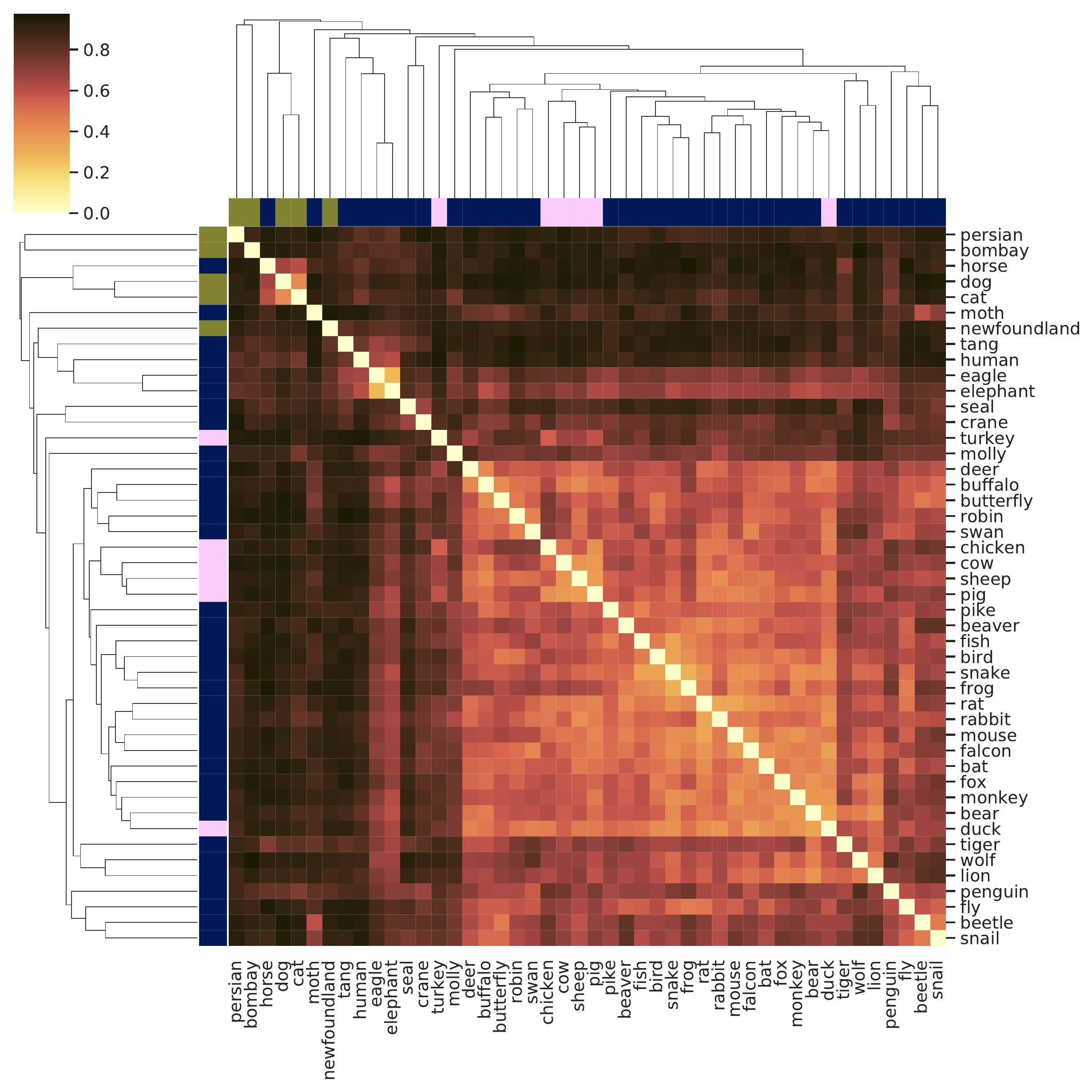}
    \caption{\reviewera{A heat map of the results of the template-based experiments, clustered by TMR with large probability changes in RoBERTa: \colorboxself{\thinpink} refers to ``farm'' animals, \colorboxself{\dirtyyellow} indicates nonhuman companions and \colorboxself{\blueblack}  stands for the remaining species. The score represents the $1-$TMR. The lighter cells, the higher similarity among animals.}}
    \label{fig:hearmap-roberta}
\end{figure}

\begin{figure}[p]
    \centering
    \includegraphics[width=0.98\linewidth]{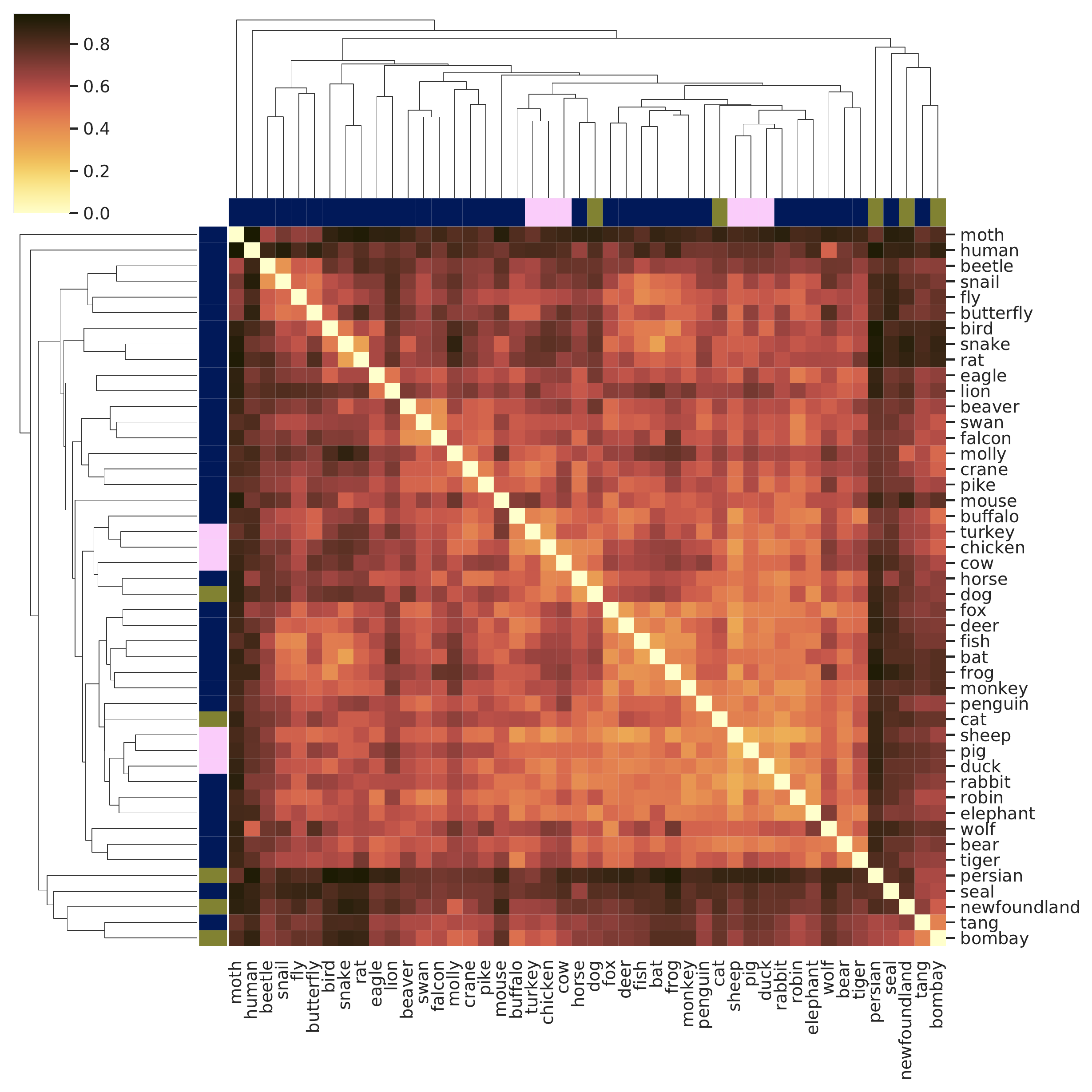}
    \caption{\reviewera{A heat map of the results of the template-based experiments, clustered by TMR with large probability changes in DistilBERT: \colorboxself{\thinpink} refers to ``farm'' animals, \colorboxself{\dirtyyellow} indicates nonhuman companions and \colorboxself{\blueblack}  stands for the remaining species. The score represents the $1-$TMR. The lighter cells, the higher similarity among animals.}}
    \label{fig:hearmap-albert}
\end{figure}

\begin{figure}[p]
    \centering
    \includegraphics[width=0.98\linewidth]{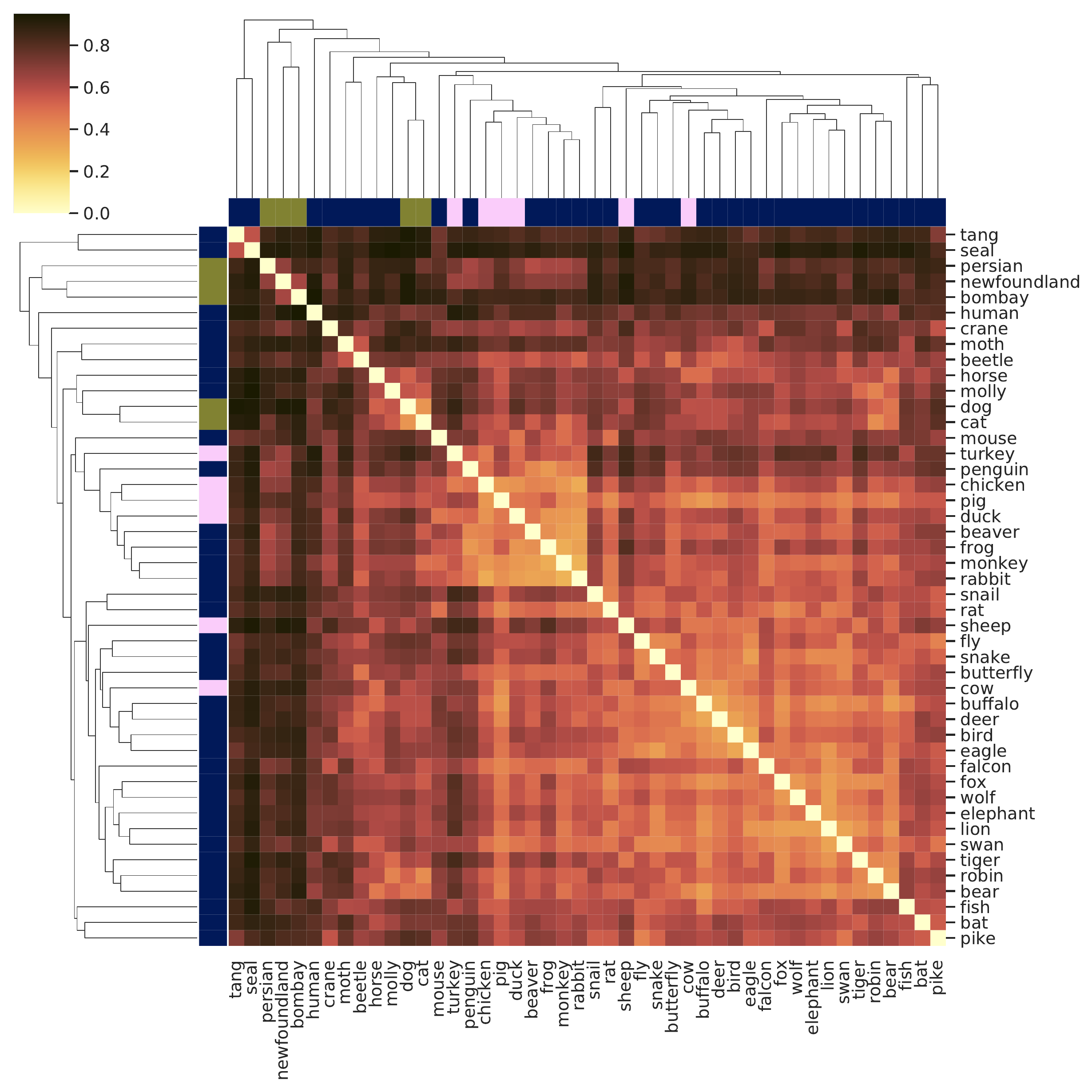}
    \caption{\reviewera{A heat map of the results of the template-based experiments, clustered by TMR with large probability changes in ALBERT: \colorboxself{\thinpink} refers to ``farm'' animals, \colorboxself{\dirtyyellow} indicates nonhuman companions and \colorboxself{\blueblack}  stands for the remaining species. The score represents the $1-$TMR. The lighter cells, the higher similarity among animals.}}
    \label{fig:hearmap-distil}
\end{figure}

\begin{figure}[ht]
    \centering
    \includegraphics[trim=20 50 50 50,clip,width=\linewidth]{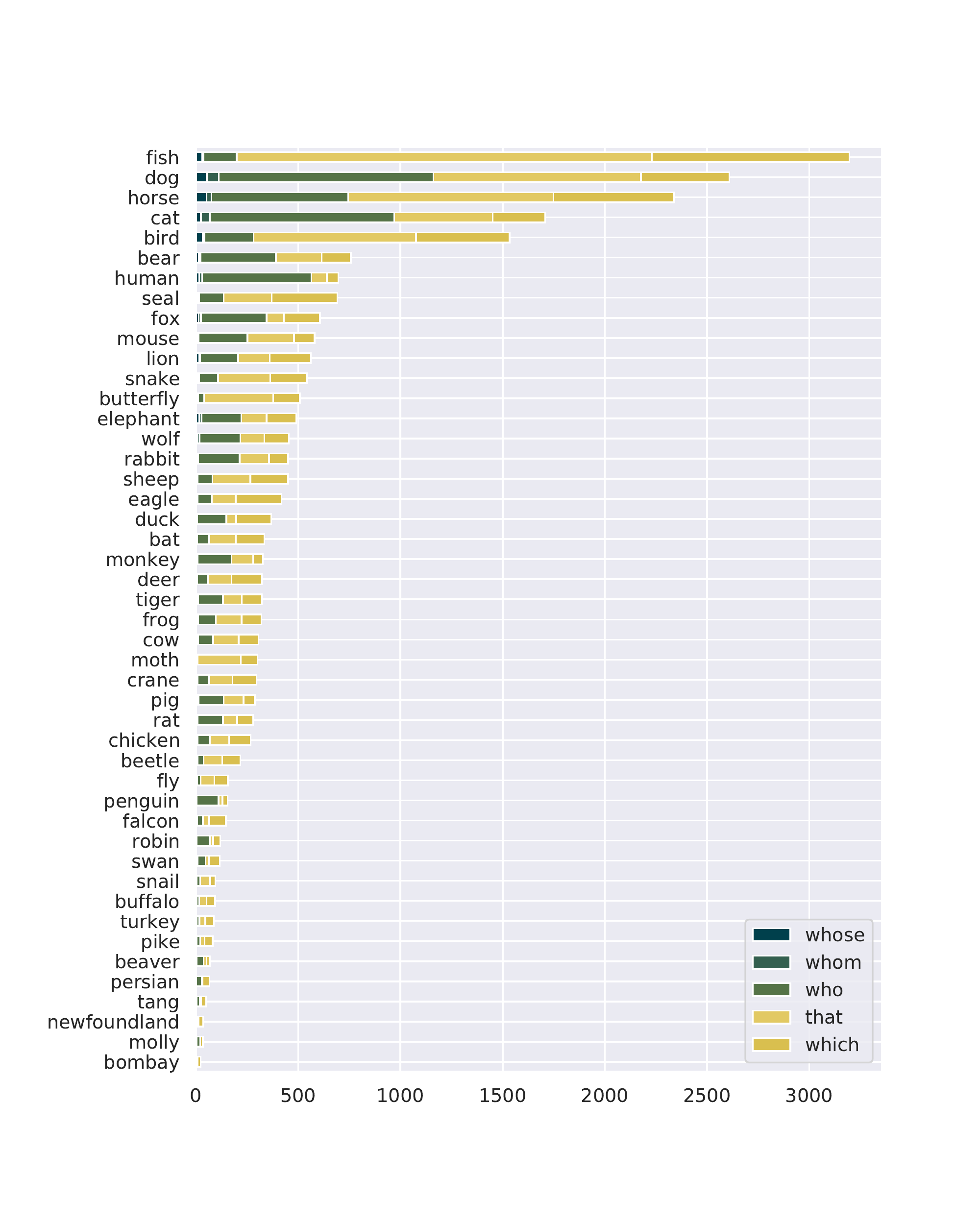}
    \caption{Number of relative pronouns referring to each animal in English Wikipedia}
    \label{fig:animal_rel_ref_freq_in_wiki}
\end{figure}

\begin{figure}[ht]
    \centering
    \includegraphics[trim=20 50 50 50,clip,width=\linewidth]{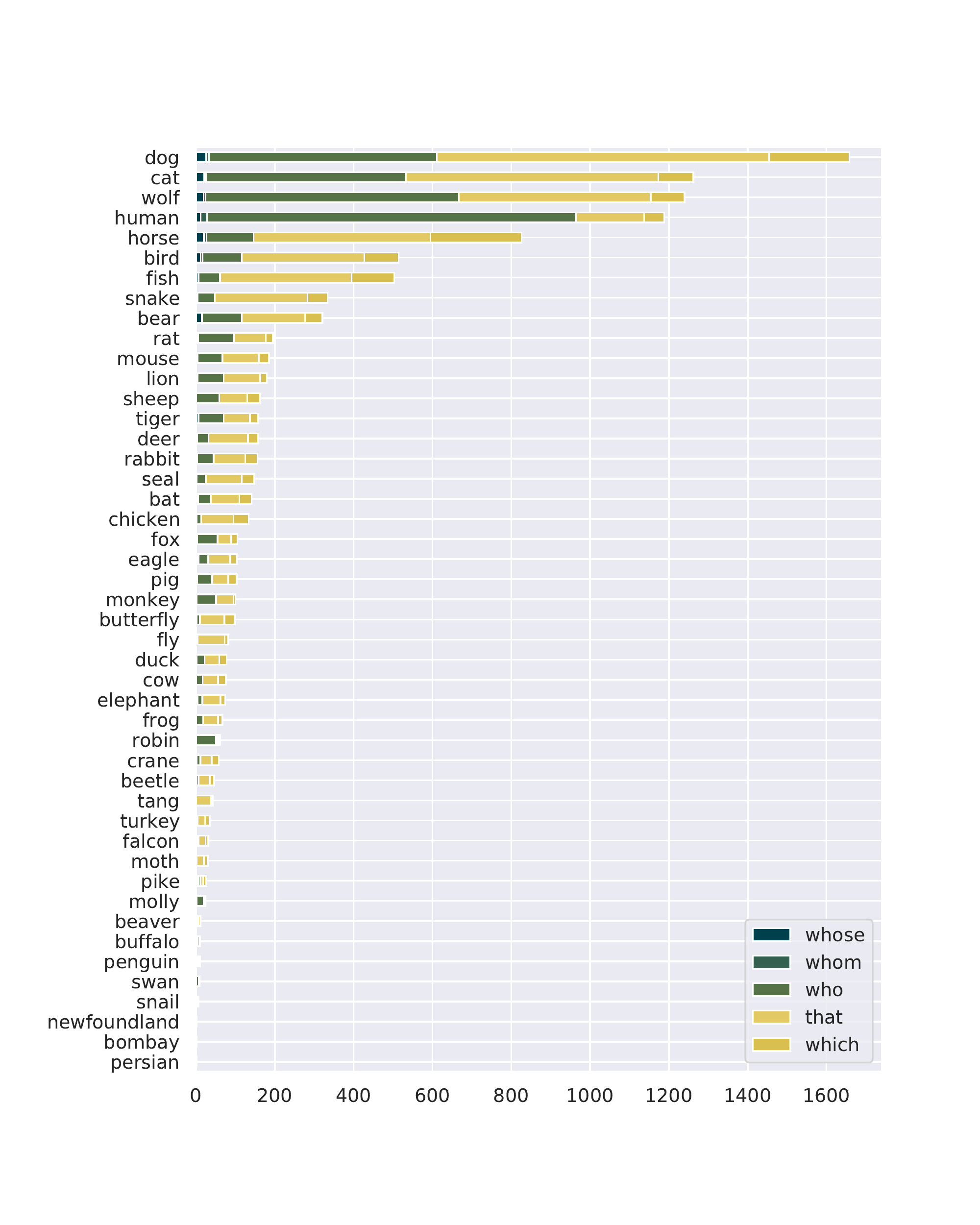}
    \caption{Number of relative pronouns referring to each animal in BookCorpus}
    \label{fig:animal_rel_ref_freq_in_bookscorpus}
\end{figure}

\begin{figure}[ht]
    \centering
    \includegraphics[trim=20 50 50 50,clip,width=\linewidth]{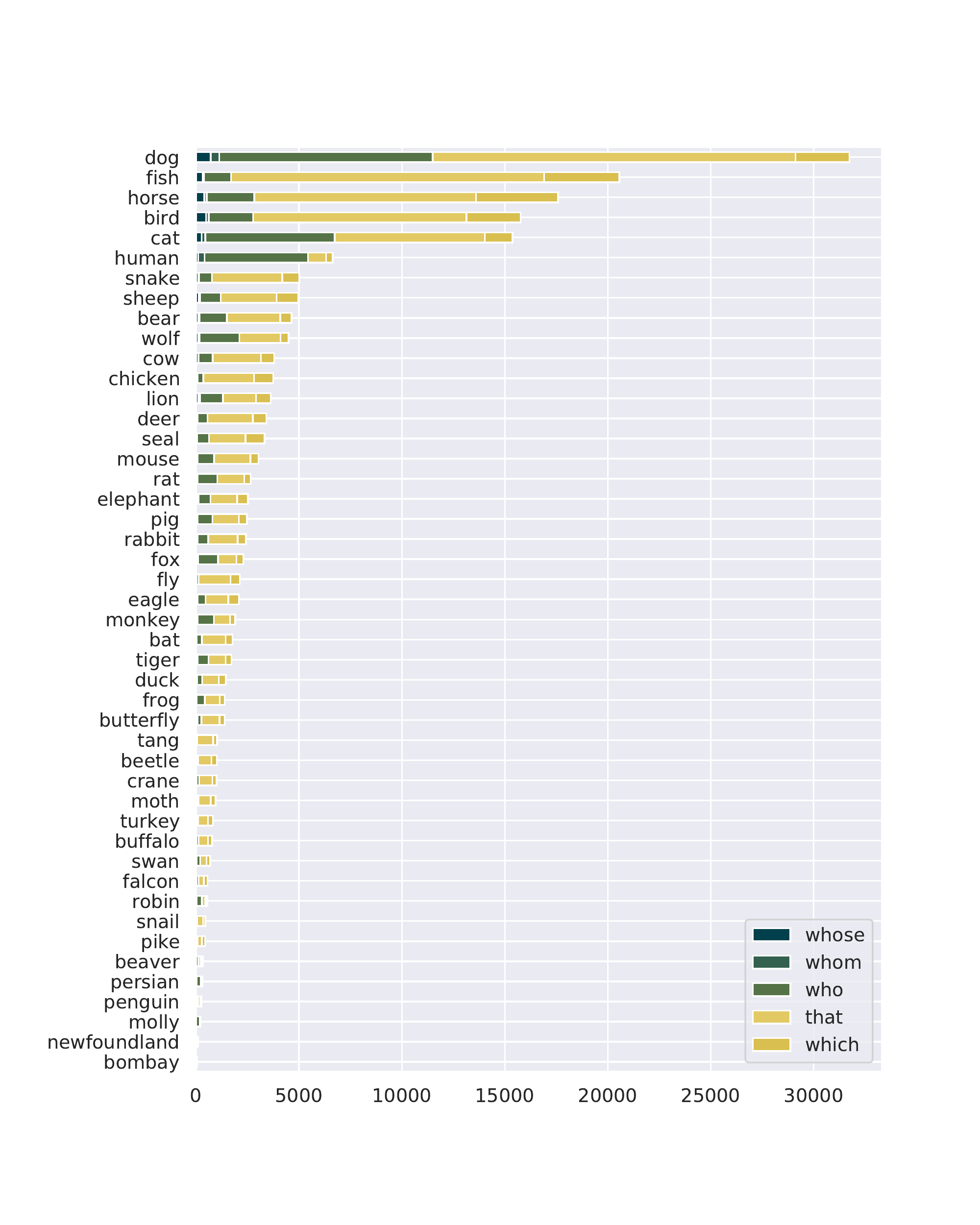}
    \caption{Number of relative pronouns referring to each animal in Books3}
    \label{fig:animal_rel_ref_freq_in_books3}
\end{figure}

\begin{table}[htb]
    \centering
    \caption{Sets of five words with the highest change rate in BERT}
    \footnotesize
    \begin{tabularx}{0.99\linewidth}{|l|X|X|}
    \hline
        animal name&  words with high probability change in \textit{object} sentences  & words with high probability change in \textit{human} sentences  \\ \hline\hline
        \colorboxself{\blueblack} bat & endemic, threatened, predatory, barred, endangered & Ninja, sarcastic, blonde, Nordic, psychic \\ \hline
        \colorboxself{\blueblack} bear & f**ked, ours, waking, happening, stirring & sarcastic, bisexual, mute, psychic, blonde \\ \hline
        \colorboxself{\blueblack} beaver & endemic, reproduced, f**ked, extinct, viable & mute, psychic, Ninja, sarcastic, superhero \\ \hline
        \colorboxself{\blueblack} beetle & conspicuous, stinging, waking, ripe, variable & coach, coaching, coaches, Swiss, midfielder \\ \hline
        \colorboxself{\blueblack} bird & endemic, threatened, uncommon, endangered, widespread & sarcastic, blonde, psychic, superhero, heroine \\ \hline
        \colorboxself{\dirtyyellow} bombay & standardized, portable, timed, ceremonial, audible & unemployed, widowed, homeless, heroine, psychologist \\ \hline
        \colorboxself{\blueblack} buffalo & stamped, f**ked, beef, slaughtered, reproduced & psychic, mute, sarcastic, clumsy, blind \\ \hline
        \colorboxself{\blueblack} butterfly & endemic, widespread, uncommon, threatened, disputed & sarcastic, superhero, blonde, cheerful, mute \\ \hline
        \colorboxself{\dirtyyellow} cat & f**ked, f**king, reproduced, violated, ripe & sarcastic, mute, Ninja, clumsy, unnamed \\ \hline
        \colorboxself{\thinpink} chicken & slaughtered, f**ked, stamped, reproduced, ripe & clumsy, mute, sarcastic, psychic, superhero \\ \hline
        \colorboxself{\thinpink} cow & f**ked, stamped, slaughter, slaughtered, ripe & sarcastic, mute, clumsy, psychic, cheerful \\ \hline
        \colorboxself{\blueblack} crane & loading, rotating, operating, overhead, tuned & psychic, blonde, sarcastic, mute, heroine \\ \hline
        \colorboxself{\blueblack} deer & beef, f**ked, f**king, viable, barred & mute, fairy, sarcastic, Ariel, psychic \\ \hline
        \colorboxself{\dirtyyellow} dog & f**ked, f**king, struck, violated, committed & sarcastic, Ninja, mute, bisexual, unnamed \\ \hline
        \colorboxself{\thinpink} duck & endemic, f**ked, reproduced, endangered, viable & sarcastic, psychic, clumsy, mute, cheerful \\ \hline
        \colorboxself{\blueblack} eagle & barred, circling, happening, endemic, yours & mute, psychic, sarcastic, Amazon, blonde \\ \hline
        \colorboxself{\blueblack} elephant & f**ked, stamped, reproduced, f**king, happening & sarcastic, mute, psychic, clumsy, cheerful \\ \hline
        \colorboxself{\blueblack} falcon & barred, endemic, f**ked, reproduced, extinct & blonde, Ninja, psychic, sarcastic, Amazon \\ \hline
        \colorboxself{\blueblack} fish & endemic, predatory, widespread, perennial, barred & heroine, sarcastic, Cinderella, princess, cheerful \\ \hline
        \colorboxself{\blueblack} fly & predatory, toxic, stinging, endemic, colonial & sarcastic, cheerful, superhero, blonde, genius \\ \hline
        \colorboxself{\blueblack} fox & f**ked, happening, waking, calling, ours & mute, sarcastic, blonde, bisexual, clumsy \\ \hline
        \colorboxself{\blueblack} frog & endemic, threatened, endangered, \#\#olate, widespread & sarcastic, Ninja, cheerful, clumsy, blonde \\ \hline
        \colorboxself{\blueblack} horse & f**ked, sin, violated, stamped, ripe & unnamed, pink, sarcastic, blonde, Ariel \\ \hline
        \colorboxself{\blueblack} human & ourselves, worth, ours, yours, our & bisexual, Ninja, sarcastic, blonde, lesbian \\ \hline
        \colorboxself{\blueblack} lion & ours, happening, waking, arising, pictured & psychic, sarcastic, mute, heroine, bisexual \\ \hline
        \colorboxself{\blueblack} molly & unacceptable, theirs, treason, happening, occurring & blonde, mute, deaf, cheerful, widowed \\ \hline
        \colorboxself{\blueblack} monkey & f**ked, f**king, waking, happening, ours & sarcastic, mute, clumsy, lesbian, Ninja \\ \hline
        \colorboxself{\blueblack} moth & endemic, Crambidae, \#\#tropical, variable, Geometridae & unemployed, Ninja, DJ, psychic, undefeated \\ \hline
        \colorboxself{\blueblack} mouse & reproduced, viable, mating, f**ked, endemic & sarcastic, cheerful, clumsy, Dorothy, superhero \\ \hline
        \colorboxself{\dirtyyellow} newfoundland & ours, theirs, happening, paradise, nearer & bisexual, protagonist, narrator, heroine, blonde \\ \hline
        \colorboxself{\blueblack} penguin & endemic, extinct, endangered, barred, reproduced & sarcastic, psychic, Ninja, clumsy, mute \\ \hline
        \colorboxself{\dirtyyellow} persian & periodic, convex, contraction, symmetric, bounded & deaf, genius, widowed, \#\#headed, intelligent \\ \hline
        \colorboxself{\thinpink} pig & f**ked, stamped, slaughtered, reproduced, sin & clumsy, sarcastic, mute, cheerful, blonde \\ \hline
        \colorboxself{\blueblack} pike & endemic, barred, preferred, edged, subspecies & blonde, mute, widowed, cheerful, homeless \\ \hline
        \colorboxself{\blueblack} rabbit & f**ked, waking, happening, slaughtered, arriving & mute, sarcastic, bisexual, psychic, clumsy \\ \hline
        \colorboxself{\blueblack} rat & reproduced, f**ked, viable, reared, waking & sarcastic, mute, clumsy, Gothic, cheerful \\ \hline
        \colorboxself{\blueblack} robin & endemic, subspecies, threatened, barred, unmistakable & mute, sarcastic, psychic, cheerful, mechanic \\ \hline
        \colorboxself{\blueblack} seal & stamped, forged, valid, void, binding & Brave, blonde, Ninja, psychic, mute \\ \hline
        \colorboxself{\thinpink} sheep & endemic, f**ked, sustainable, perennial, viable & mute, sarcastic, psychic, princess, narrator \\ \hline
        \colorboxself{\blueblack} snail & predatory, endemic, widespread, fossil, marine & sarcastic, cheerful, mute, optimistic, psychic \\ \hline
        \colorboxself{\blueblack} snake & endemic, yours, barred, ours, venom & sarcastic, cheerful, blonde, mute, optimistic \\ \hline
        \colorboxself{\blueblack} swan & yours, ours, f**ked, reproduced, endemic & psychic, sarcastic, mute, mechanic, clumsy \\ \hline
        \colorboxself{\blueblack} tang & audible, repeated, nasal, consonant, pronounced & Smart, unemployed, smart, homeless, brave \\ \hline
        \colorboxself{\blueblack} tiger & happening, ours, f**king, waking, f**ked & mute, sarcastic, psychic, bisexual, blonde \\ \hline
        \colorboxself{\thinpink} turkey & stamped, slaughtered, beef, ripe, viable & mute, clumsy, psychic, sarcastic, deaf \\ \hline
        \colorboxself{\blueblack} wolf & ours, yours, happening, waking, you & bisexual, mute, sarcastic, psychic, lesbian \\ \hline
    \end{tabularx}
\end{table}

\begin{table}[p]
    \centering
    \caption{Sets of five words with the highest change rate in RoBERTa}
    \footnotesize
    \begin{tabularx}{0.99\linewidth}{|l|X|X|}
    \hline
       animal name&  words with high probability change in \textit{object} sentences  & words with high probability change in \textit{human} sentences  \\ \hline\hline
        \colorboxself{\blueblack} bat &  intact,  handled,  dried,  unloaded,  batted &  virtuous,  heroic,  witty,  superhuman,  princess \\ \hline
        \colorboxself{\blueblack} bear &  polled,  extant,  freshwater,  handled,  endemic &  superhuman,  mercenary,  romantic,  sarcastic,  prince \\ \hline
        \colorboxself{\blueblack} beaver &  invasive,  freshwater,  widespread,  dried,  common &  atheist,  lonely,  swearing,  nineteen,  lesbian \\ \hline
        \colorboxself{\blueblack} beetle &  deposited,  feeding,  circulating,  hardest,  clustered &  virtuous,  heroic,  fictional,  philosophical,  courageous \\ \hline
        \colorboxself{\blueblack} bird &  freshwater,  offshore,  migr,  endemic,  extant &  Human,  philosophical,  jealous,  sarcastic,  witty \\ \hline
        \colorboxself{\dirtyyellow} bombay &  fallacy,  phosphorus,  absurdity,  gelatin,  FALSE &  Shy,  loyal,  married,  shy,  wealthy \\ \hline
        \colorboxself{\blueblack} buffalo &  dried,  freshwater,  listed,  polled,  stamped &  cowardly,  arrogant,  selfish,  cunning,  rebellious \\ \hline
        \colorboxself{\blueblack} butterfly &  variable,  common,  offshore,  widespread,  clustered &  virtuous,  superhuman,  philosophical,  rebellious,  heroic \\ \hline
        \colorboxself{\dirtyyellow} cat &  terrestrial,  armoured,  netted,  scaled,  predatory &  foster,  deaf,  Transgender,  Blind,  Polish \\ \hline
        \colorboxself{\thinpink} chicken &  dried,  freshwater,  semen,  polled,  harvested &  optimistic,  sarcastic,  romantic,  pessimistic,  Psychic \\ \hline
        \colorboxself{\thinpink} cow &  polled,  dried,  semen,  domestically,  processed &  romantic,  optimistic,  witty,  poetic,  mysterious \\ \hline
        \colorboxself{\blueblack} crane &  erected,  automated,  propelled,  loader,  towed &  jealous,  psychic,  horny,  deaf,  conflicted \\ \hline
        \colorboxself{\blueblack} deer &  bucks,  dried,  harvested,  roadside,  buck &  swearing,  witty,  romantic,  philosophical,  jealous \\ \hline
        \colorboxself{\dirtyyellow} dog &  terrestrial,  itself,  predatory,  defined,  armoured &  deaf,  transsexual,  foster,  Homeless,  lesbian \\ \hline
        \colorboxself{\thinpink} duck &  freshwater,  polled,  dried,  offshore,  netted &  superhuman,  heroic,  superhero,  protagonist,  Human \\ \hline
        \colorboxself{\blueblack} eagle &  correlated,  achievable,  warranted,  measurable,  irreversible &  adventurer,  hacker,  Paladin,  Sailor,  trainer \\ \hline
        \colorboxself{\blueblack} elephant &  achievable,  warranted,  happening,  extinct,  irreversible &  adventurer,  detective,  Lesbian,  thief,  vigilante \\ \hline
        \colorboxself{\blueblack} falcon &  freshwater,  netted,  largest,  aerial,  perched &  Human,  optimistic,  superhuman,  rebellious,  lesbian \\ \hline
        \colorboxself{\blueblack} fish &  freshwater,  reef,  widespread,  polled,  aggregate &  swearing,  jealous,  witty,  superhuman,  sixteen \\ \hline
        \colorboxself{\blueblack} fly &  respiratory,  common,  genital,  dried,  larvae &  heroic,  lonely,  witty,  Talking,  intuitive \\ \hline
        \colorboxself{\blueblack} fox &  polled,  invasive,  Madagascar,  pictured,  extant &  pessimistic,  sarcastic,  mercenary,  romantic,  compassionate \\ \hline
        \colorboxself{\blueblack} frog &  freshwater,  larvae,  widespread,  invasive,  dart &  superhuman,  seventeen,  nineteen,  swearing,  heroic \\ \hline
        \colorboxself{\blueblack} horse &  clicking,  enough,  beat,  it,  right &  Transgender,  lesbian,  deaf,  transgender,  transsexual \\ \hline
        \colorboxself{\blueblack} human &  extant,  extinct,  ours,  yours,  edible &  bartender,  nineteen,  seventeen,  sixteen,  eighteen \\ \hline
        \colorboxself{\blueblack} lion &  pictured,  Madagascar,  Guinea,  polled,  Bengal &  Human,  prince,  princess,  mercenary,  Princess \\ \hline
        \colorboxself{\blueblack} molly &  edible,  larvae,  harvested,  dried,  invasive &  pessimistic,  Persian,  nineteen,  deaf,  lazy \\ \hline
        \colorboxself{\blueblack} monkey &  polled,  palm,  Madagascar,  extant,  Guinea &  virtuous,  mercenary,  superhuman,  Alone,  romantic \\ \hline
        \colorboxself{\blueblack} moth &  happening,  circulating,  newer,  collapsing,  getting &  prophetic,  divine,  :,  Blind,  feminist \\ \hline
        \colorboxself{\blueblack} mouse &  polled,  larvae,  extant,  freshwater,  edible &  swearing,  romantic,  Alone,  heroic,  rich \\ \hline
        \colorboxself{\dirtyyellow} newfoundland &  Antarctica,  unfolding,  contiguous,  ours,  wetlands &  deaf,  transsexual,  bisexual,  runner,  addicted \\ \hline
        \colorboxself{\blueblack} penguin &  lower,  offshore,  flattened,  freshwater,  oval &  lesbian,  unmarried,  married,  rebellious,  feminist \\ \hline
        \colorboxself{\dirtyyellow} persian &  larvae,  edible,  peeled,  citrus,  vegetation &  atheist,  writer,  novelist,  journalist,  physicist \\ \hline
        \colorboxself{\thinpink} pig &  polled,  dried,  harvested,  yielded,  peeled &  romantic,  selfish,  optimistic,  jealous,  arrogant \\ \hline
        \colorboxself{\blueblack} pike &  freshwater,  offshore,  invasive,  Atlantic,  harvested &  Human,  protector,  nineteen,  optimistic,  swearing \\ \hline
        \colorboxself{\blueblack} rabbit &  dried,  widespread,  netted,  terrestrial,  harvested &  sarcastic,  Psychic,  optimistic,  pessimistic,  heroic \\ \hline
        \colorboxself{\blueblack} rat &  dried,  freshwater,  widespread,  polled,  extant &  heroic,  swearing,  superhuman,  romantic,  protector \\ \hline
        \colorboxself{\blueblack} robin &  common,  variable,  migrating,  widespread,  larvae &  superhuman,  virtuous,  philosophical,  trustworthy,  irresponsible \\ \hline
        \colorboxself{\blueblack} seal &  tightening,  tightened,  tighter,  stamped,  dried &  autistic,  Hungry,  dreaming,  deaf,  transsexual \\ \hline
        \colorboxself{\thinpink} sheep &  polled,  dried,  harvested,  yielded,  processed &  jealous,  witty,  arrogant,  heroic,  optimistic \\ \hline
        \colorboxself{\blueblack} snail &  minute,  deposited,  dried,  flattened,  occurring &  clueless,  Psychic,  jealous,  cowardly,  loyal \\ \hline
        \colorboxself{\blueblack} snake &  freshwater,  netted,  dried,  invasive,  widespread &  superhuman,  swearing,  cursed,  immortal,  protagonist \\ \hline
        \colorboxself{\blueblack} swan &  freshwater,  aerial,  lower,  largest,  netted &  protector,  trustworthy,  forgiving,  pessimistic,  loyal \\ \hline
        \colorboxself{\blueblack} tang &  contraction,  residue,  correlation,  causation,  correlated &  deaf,  homeless,  transsexual,  Homeless,  veterinarian \\ \hline
        \colorboxself{\blueblack} tiger &  manageable,  corrected,  viable,  right,  largest &  lesbian,  princess,  transsexual,  vegan,  Human \\ \hline
        \colorboxself{\thinpink} turkey &  dried,  processed,  ground,  slaughtered,  cached &  deaf,  listening,  jealous,  optimistic,  psychic \\ \hline
        \colorboxself{\blueblack} wolf &  polled,  extant,  heaviest,  widespread,  invasive &  Psychic,  wizard,  Human,  Loki,  prince \\ \hline
    \end{tabularx}
\end{table}

\begin{table}[p]
    \centering
    \caption{Sets of five words with the highest change rate in DistilBERT}
    \footnotesize
    \begin{tabularx}{0.99\linewidth}{|l|X|X|}
    \hline
        animal name&  words with high probability change in \textit{object} sentences  & words with high probability change in \textit{human} sentences  \\ \hline\hline
        \colorboxself{\blueblack} bat & endemic, distributed, widespread, \#\#olate, \#\#gratory & magician, psychic, witch, villains, wizard \\ \hline
        \colorboxself{\blueblack} bear & endemic, distributed, valid, edible, convex & psychic, witches, witch, herself, grandmother \\ \hline
        \colorboxself{\blueblack} beaver & distributed, endemic, lateral, \#\#gratory, inactivated & psychic, heroine, archaeologist, magician, narrator \\ \hline
        \colorboxself{\blueblack} beetle & endemic, widespread, distributed, subsp, valid & magician, transgender, psychic, widowed, deaf \\ \hline
        \colorboxself{\blueblack} bird & endemic, distributed, widespread, variable, declining & robot, psychic, princess, witches, angel \\ \hline
        \colorboxself{\dirtyyellow} bombay & quarterly, annual, administered, recited, yearly & widowed, transgender, deaf, bisexual, blind \\ \hline
        \colorboxself{\blueblack} buffalo & endemic, abolished, extinct, inactivated, edible & heroine, actress, girlfriend, psychic, narrator \\ \hline
        \colorboxself{\blueblack} butterfly & endemic, widespread, distributed, valid, decreasing & lion, psychic, controlling, gifted, vain \\ \hline
        \colorboxself{\dirtyyellow} cat & endemic, valid, convex, inactivated, viable & narrator, thirteen, psychic, fourteen, seventeen \\ \hline
        \colorboxself{\thinpink} chicken & endemic, edible, pounded, differentiated, clarified & deaf, blind, psychic, narrator, bullying \\ \hline
        \colorboxself{\thinpink} cow & endemic, edible, differentiated, sacred, branched & homeless, deaf, bullying, blind, paranoid \\ \hline
        \colorboxself{\blueblack} crane & distributed, towed, valid, endemic, unfolded & psychic, heroine, deaf, magician, actress \\ \hline
        \colorboxself{\blueblack} deer & endemic, \#\#gratory, distributed, extinct, subspecies & psychic, sailor, narrator, witch, grandmother \\ \hline
        \colorboxself{\dirtyyellow} dog & endemic, subspecies, branched, differentiated, valid & herself, teenage, widowed, thirteen, grandmother \\ \hline
        \colorboxself{\thinpink} duck & endemic, valid, distributed, subspecies, edible & psychic, narrator, clumsy, deaf, thirteen \\ \hline
        \colorboxself{\blueblack} eagle & endemic, distributed, valid, lateral, decreasing & psychic, princess, witches, herself, fairies \\ \hline
        \colorboxself{\blueblack} elephant & endemic, distributed, convex, valid, inhabited & heroine, magician, nurse, psychic, princess \\ \hline
        \colorboxself{\blueblack} falcon & convex, scaled, lateral, distributed, endemic & psychic, transgender, magician, controlling, kidnapped \\ \hline
        \colorboxself{\blueblack} fish & endemic, distributed, widespread, variable, diagnostic & widowed, narrator, sailor, genius, girlfriend \\ \hline
        \colorboxself{\blueblack} fly & distributed, valid, extant, endemic, occurring & deaf, motorcycle, sailor, narrator, thirteen \\ \hline
        \colorboxself{\blueblack} fox & endemic, distributed, \#\#gratory, extinct, extant & heroine, magician, psychic, sailor, narrator \\ \hline
        \colorboxself{\blueblack} frog & endemic, distributed, widespread, variable, valid & vain, princess, fairies, psychic, narrator \\ \hline
        \colorboxself{\blueblack} horse & valid, equivalent, propelled, endemic, assessed & heroine, grandmother, witches, fairies, princess \\ \hline
        \colorboxself{\blueblack} human & worth, acceptable, our, reproduced, valid & princess, witch, emerald, angel, witches \\ \hline
        \colorboxself{\blueblack} lion & endemic, engraved, displayed, seated, valid & psychic, heroine, princess, witches, witch \\ \hline
        \colorboxself{\blueblack} molly & frequented, underway, inhabited, unfinished, excavated & bisexual, deaf, transgender, elderly, widowed \\ \hline
        \colorboxself{\blueblack} monkey & endemic, distributed, valid, convex, differentiated & princess, witch, magician, herself, witches \\ \hline
        \colorboxself{\blueblack} moth & widespread, occurring, varies, irregular, subsp & blind, sighted, deaf, blinded, astronomer \\ \hline
        \colorboxself{\blueblack} mouse & distributed, endemic, inactivated, valid, bilateral & witches, fairies, thirteen, prostitutes, witch \\ \hline
        \colorboxself{\dirtyyellow} newfoundland & endemic, populated, inhabited, frequented, dotted & widowed, secretary, bisexual, transgender, pregnant \\ \hline
        \colorboxself{\blueblack} penguin & endemic, valid, distributed, extinct, extant & psychic, widowed, narrator, magician, actress \\ \hline
        \colorboxself{\dirtyyellow} persian & convex, bounded, periodic, continuous, compact & transgender, actress, widowed, nurse, wrestler \\ \hline
        \colorboxself{\thinpink} pig & endemic, edible, viable, differentiated, inactivated & thirteen, dolls, narrator, girlfriend, seventeen \\ \hline
        \colorboxself{\blueblack} pike & valid, longitudinal, endemic, distributed, convex & psychic, heroine, fairies, caring, narrator \\ \hline
        \colorboxself{\blueblack} rabbit & endemic, distributed, viable, differentiated, inactivated & magician, psychic, witch, witches, narrator \\ \hline
        \colorboxself{\blueblack} rat & endemic, distributed, oral, lateral, bilateral & witches, witch, fairies, wizard, princess \\ \hline
        \colorboxself{\blueblack} robin & endemic, distributed, valid, branched, widespread & psychic, heroine, autism, deaf, narrator \\ \hline
        \colorboxself{\blueblack} seal & stamped, filed, valid, worn, engraved & heroine, psychic, kidnapped, protagonist, drowning \\ \hline
        \colorboxself{\thinpink} sheep & endemic, distributed, inactivated, viable, extinct & psychic, housekeeper, narrator, thirteen, witch \\ \hline
        \colorboxself{\blueblack} snail & widespread, distributed, endemic, variable, minute & psychic, villain, widowed, protagonist, lion \\ \hline
        \colorboxself{\blueblack} snake & distributed, endemic, \#\#olate, variable, diagnostic & witch, princess, fairies, wizard, goddess \\ \hline
        \colorboxself{\blueblack} swan & endemic, distributed, \#\#tail, lateral, \#\#gratory & psychic, narrator, magician, transgender, autism \\ \hline
        \colorboxself{\blueblack} tang & recited, oral, cumulative, meaningful, elastic & blind, widowed, heroine, deaf, scientist \\ \hline
        \colorboxself{\blueblack} tiger & endemic, distributed, valid, extant, inhabited & heroine, widowed, princess, lovers, witch \\ \hline
        \colorboxself{\thinpink} turkey & endemic, extant, edible, widespread, valid & deaf, actress, psychic, transgender, narrator \\ \hline
        \colorboxself{\blueblack} wolf & endemic, valid, conspicuous, edible, variable & witches, witch, princess, fairies, grandmother \\ \hline
    \end{tabularx}
\end{table}

\begin{table*}[htbp]
    \centering
    \caption{Sets of five words with the highest change rate in ALBERT}
    \footnotesize
    \begin{tabularx}{0.99\linewidth}{|l|X|X|}
    \hline
        animal name&  words with high probability change in \textit{object} sentences  & words with high probability change in \textit{human} sentences  \\ \hline\hline
        \colorboxself{\blueblack} bat & printed,  basalt,  lodged,  cylindrical,  mandible  & confident,  gambler,  dreamer,  fearless,  grieving  \\ \hline
        \colorboxself{\blueblack} bear & lodged,  reported,  indicated,  suggested,  excavated  & trusting,  helpless,  fearless,  trusted,  obedient  \\ \hline
        \colorboxself{\blueblack} beaver & noticeable,  lodged,  brownish,  coughed,  yellowish  & heiress,  dreamer,  princess,  bachelor,  addict  \\ \hline
        \colorboxself{\blueblack} beetle & leaked,  brownish,  lodged,  occurring,  yellowish  & princess,  adventurer,  dreamer,  valkyrie,  knighted  \\ \hline
        \colorboxself{\blueblack} bird & printed,  brownish,  yellowish,  lodged,  localized  & dreamer,  conqueror,  princess,  angels,  slaves  \\ \hline
        \colorboxself{\dirtyyellow} bombay & reopened,  commenced,  redeveloped,  expanded,  skyline  & widow,  soprano,  eunuch,  knighted,  pregnant  \\ \hline
        \colorboxself{\blueblack} buffalo & brownish,  lodged,  yellowish,  reported,  basalt  & dreamer,  hero,  helpless,  obedient,  widow  \\ \hline
        \colorboxself{\blueblack} butterfly & printed,  yellowish,  brownish,  highlighted,  forewing  & dreamer,  conqueror,  adventurer,  himself,  superhuman  \\ \hline
        \colorboxself{\dirtyyellow} cat & lodged,  boar,  appeared,  urine,  yellowish  & dreamer,  fearless,  bachelor,  confident,  jed  \\ \hline
        \colorboxself{\thinpink} chicken & spelt,  brownish,  lodged,  compressed,  stemmed  & dreamer,  atheist,  jealous,  telepathic,  princess  \\ \hline
        \colorboxself{\thinpink} cow & weigh,  spelt,  raked,  brownish,  lodged  & destiny,  conqueror,  happiness,  trusting,  fearless  \\ \hline
        \colorboxself{\blueblack} crane & corrugated,  aluminium,  hangar,  diameter,  turbine  & jealous,  dreamer,  eunuch,  homosexual,  tigre  \\ \hline
        \colorboxself{\blueblack} deer & brownish,  lodged,  reported,  yellowish,  surfaced  & dreamer,  slaves,  conqueror,  trusting,  angels  \\ \hline
        \colorboxself{\dirtyyellow} dog & lodged,  boar,  suggested,  reported,  spelt  & perfection,  caring,  fearless,  loving,  faithful  \\ \hline
        \colorboxself{\thinpink} duck & spelt,  contains,  termed,  spelled,  containing  & atheist,  adventurer,  dreamer,  addict,  estranged  \\ \hline
        \colorboxself{\blueblack} eagle & resembled,  printed,  brachy,  tapered,  holotype  & conqueror,  helpless,  widow,  steward,  dreamer  \\ \hline
        \colorboxself{\blueblack} elephant & reported,  brownish,  yellowish,  lodged,  surfaced  & helpless,  conqueror,  obedient,  slaves,  estranged  \\ \hline
        \colorboxself{\blueblack} falcon & resembled,  compressed,  mandible,  resembles,  rectangular  & dreamer,  fearless,  trusting,  addict,  obedient  \\ \hline
        \colorboxself{\blueblack} fish & tapered,  formulated,  brownish,  stemmed,  tasted  & dreamer,  jealous,  himself,  atheist,  conqueror  \\ \hline
        \colorboxself{\blueblack} fly & nitrogen,  printed,  tapered,  compressed,  brownish  & conqueror,  dreamer,  hostage,  slaves,  murderer  \\ \hline
        \colorboxself{\blueblack} fox & brownish,  yellowish,  dorsal,  bluish,  puma  & dreamer,  trusted,  slaves,  trusting,  selfish  \\ \hline
        \colorboxself{\blueblack} frog & resembled,  contains,  spelt,  termed,  compressed  & atheist,  princess,  bachelor,  transgender,  dreamer  \\ \hline
        \colorboxself{\blueblack} horse & hoof,  suggested,  raked,  overturned,  hydraulic  & caring,  fearless,  helpless,  trusting,  perfection  \\ \hline
        \colorboxself{\blueblack} human & http,  suggested,  computed,  spelt,  stated  & savior,  loves,  protector,  loving,  beloved  \\ \hline
        \colorboxself{\blueblack} lion & noticeable,  indicated,  reported,  yellowish,  conical  & trusting,  estranged,  helpless,  dreamer,  selfish  \\ \hline
        \colorboxself{\blueblack} molly & spelled,  suggested,  advertised,  yellowish,  bacterio  & confidant,  dreamer,  confident,  obedient,  fearless  \\ \hline
        \colorboxself{\blueblack} monkey & resembled,  spelt,  xylo,  termed,  suggested  & estranged,  princess,  dreamer,  atheist,  bachelor  \\ \hline
        \colorboxself{\blueblack} moth & widespread,  annual,  biennial,  localized,  basal  & dreamer,  jed,  magician,  sorcerer,  himself  \\ \hline
        \colorboxself{\blueblack} mouse & generate,  termed,  contains,  kernel,  xml  & wealthy,  fearless,  princess,  billionaire,  dreamer  \\ \hline
        \colorboxself{\dirtyyellow} newfoundland & happen,  happened,  place,  reopened,  resumed  & widow,  knighted,  pregnant,  transgender,  addict  \\ \hline
        \colorboxself{\blueblack} penguin & contained,  brownish,  noticeable,  smelled,  yellowish  & widow,  atheist,  addict,  billionaire,  heiress  \\ \hline
        \colorboxself{\dirtyyellow} persian & quartz,  sodium,  clarified,  indicated,  contrary  & heiress,  wealthy,  married,  widow,  unmarried  \\ \hline
        \colorboxself{\thinpink} pig & brownish,  termed,  dorsal,  yellowish,  spelt  & trusting,  jealous,  princess,  selfish,  estranged  \\ \hline
        \colorboxself{\blueblack} pike & corrugated,  diameter,  tapered,  aluminium,  compressed  & jealous,  gambler,  grieving,  helpless,  dreamer  \\ \hline
        \colorboxself{\blueblack} rabbit & snout,  resembled,  contains,  termed,  spelt  & dreamer,  atheist,  princess,  transgender,  estranged  \\ \hline
        \colorboxself{\blueblack} rat & nitrogen,  termed,  contains,  1:,  containing  & dreamer,  selfish,  conqueror,  estranged,  atheist  \\ \hline
        \colorboxself{\blueblack} robin & plumage,  brownish,  yellowish,  printed,  spelt  & dreamer,  confident,  heroine,  psychopath,  selfish  \\ \hline
        \colorboxself{\blueblack} seal & minimize,  compress,  tissue,  membrane,  corrugated  & temeraire,  racehorse,  knighted,  valkyrie,  shepherd  \\ \hline
        \colorboxself{\thinpink} sheep & aerobic,  discontinued,  uploaded,  dorsal,  reported  & traitor,  helpless,  trusting,  conqueror,  slaves  \\ \hline
        \colorboxself{\blueblack} snail & nitrogen,  termed,  corrugated,  sodium,  containing  & selfish,  dreamer,  strangers,  helpless,  obedient  \\ \hline
        \colorboxself{\blueblack} snake & localized,  bluish,  yellowish,  pointed,  brownish  & messiah,  conqueror,  dreamer,  helpless,  obedient  \\ \hline
        \colorboxself{\blueblack} swan & printed,  tapered,  erupted,  conical,  plumage  & helpless,  obedient,  trusting,  conqueror,  estranged  \\ \hline
        \colorboxself{\blueblack} tang & nitrogen,  minimize,  termed,  compressed,  compress  & adventurer,  conqueror,  abbess,  empress,  barbarian  \\ \hline
        \colorboxself{\blueblack} tiger & yellowish,  brownish,  bluish,  excavated,  reported  & helpless,  trusting,  selfish,  caretaker,  incapable  \\ \hline
        \colorboxself{\thinpink} turkey & tasted,  dried,  sliced,  crisp,  highlighted  & adventurer,  atheist,  transgender,  telepathic,  knighted  \\ \hline
        \colorboxself{\blueblack} wolf & mandible,  dorsal,  conical,  termed,  brownish  & dreamer,  helpless,  princess,  orphan,  traitor  \\ \hline
    \end{tabularx}
\end{table*}

\begin{figure}[htb]
    \centering
    \scalebox{0.9}{
    \begin{tabular}{c}
        \begin{minipage}{0.98\linewidth}
         \centering
         \includegraphics[trim=100 25 95 20, clip,width=\linewidth]{figures/mask_prob_bias_in_Book3_bertl.pdf}
         \subcaption{BERT}
         \label{fig:bias-bert-inappendix}
        \end{minipage} \\
        \begin{minipage}{0.98\linewidth}
         \centering
         \includegraphics[trim=100 25 95 20, clip,width=\linewidth]{figures/mask_prob_bias_in_Book3_robertal.pdf}
         \subcaption{RoBERTa}
         \label{fig:bias-roberta-inappendix}
        \end{minipage}\\
        \begin{minipage}{0.98\linewidth}
         \centering
         \includegraphics[trim=100 25 95 20, clip,width=\linewidth]{figures/mask_prob_bias_in_Book3_distil.pdf}
         \subcaption{DistilBERT}
         \label{fig:bias-distil-inappendix}
        \end{minipage} 
          \\
        \begin{minipage}{0.98\linewidth}
         \centering
         \includegraphics[trim=100 25 95 20, clip,width=\linewidth]{figures/mask_prob_bias_in_Book3_albertl.pdf}
         \subcaption{ALBERT}
         \label{fig:bias-albert-inappendix}
        \end{minipage}
    \end{tabular}
    }
    \caption{Results of the corpus-based bias analysis, sorted by the magnitude of the bias represented by Equation \ref{eq:bias_relative_pred}. Vertical axis shows the magnitude of the bias, where positive values indicate that MLMs incorrectly insert ``that'' or ``which'', and negative values indicate that MLMs incorrectly insert ``who'', ``whose'', or ``whom'' with higher probability. The horizontal one shows the animal names.
    }
    \label{fig:biasinbook3-inappendix}
\end{figure} 





\end{document}